\documentclass[10pt,twocolumn,letterpaper]{article}

\usepackage{cvpr}              

\usepackage{graphicx}
\usepackage{amsmath}
\usepackage{amssymb}
\usepackage{amsthm}
\usepackage{booktabs}
\usepackage{dsfont}

\usepackage{overpic}
\usepackage{multirow}
\usepackage{enumitem} 
\usepackage{overpic} 
\usepackage{color}

\definecolor{turquoise}{cmyk}{0.65,0,0.1,0.3}
\definecolor{purple}{rgb}{0.65,0,0.65}
\definecolor{dark_green}{rgb}{0, 0.5, 0}
\definecolor{orange}{rgb}{0.8, 0.6, 0.2}
\definecolor{red}{rgb}{0.8, 0.2, 0.2}
\definecolor{darkred}{rgb}{0.6, 0.1, 0.05}
\definecolor{blueish}{rgb}{0.0, 0.3, .6}
\definecolor{light_gray}{rgb}{0.7, 0.7, .7}
\definecolor{pink}{rgb}{1, 0, 1}
\definecolor{greyblue}{rgb}{0.25, 0.25, 1}

\usepackage{blindtext}

\renewcommand{\paragraph}[1]{\vspace{1em}\noindent\textbf{#1}.}

\usepackage[pagebackref,breaklinks,colorlinks]{hyperref}

\usepackage[capitalize]{cleveref}
\crefname{section}{Sec.}{Secs.}
\Crefname{section}{Section}{Sections}
\Crefname{table}{Table}{Tables}
\crefname{table}{Tab.}{Tabs.}
\newtheorem{thm}{Theorem}
\newtheorem{cor}{Corollary}

\usepackage{amsmath,amsfonts,bm,amssymb}

\def\eqref#1{equation~\ref{#1}}

\def\1{\bm{1}}

\def\vb{{\bm{b}}}

\def\vu{{\bm{u}}}

\def\vx{{\bm{x}}}

\def\vz{{\bm{z}}}

\def\mA{{\bm{A}}}

\def\mG{{\bm{G}}}

\def\mJ{{\bm{J}}}

\def\mV{{\bm{V}}}
\def\mW{{\bm{W}}}

\DeclareMathAlphabet{\mathsfit}{\encodingdefault}{\sfdefault}{m}{sl}
\SetMathAlphabet{\mathsfit}{bold}{\encodingdefault}{\sfdefault}{bx}{n}

\def\gR{{\mathcal{R}}}
\def\gS{{\mathcal{S}}}

\def\gZ{{\mathcal{Z}}}

\DeclareMathOperator*{\argmax}{arg\,max}
\DeclareMathOperator*{\argmin}{arg\,min}

\usepackage{algorithm}
\usepackage[noend]{algpseudocode}

\def\cvprPaperID{11253} 
\def\confName{CVPR}
\def\confYear{2022}

\begin{document}
\title{
 Polarity Sampling:  Quality and Diversity Control \\ of Pre-Trained Generative Networks via Singular Values
}

\author{Ahmed Imtiaz Humayun $^{\dagger *}$, Randall Balestriero $^{\ddagger *}$, Richard Baraniuk$^\dagger$\\
\textit{Rice University{$^\dagger$}, Meta AI Research{$^\ddagger$}}
}
\maketitle
\begin{abstract}\noindent
We present Polarity Sampling, a theoretically justified plug-and-play method for controlling the generation quality and diversity of any pre-trained deep generative network (DGN).
Leveraging the fact that DGNs are, or can be approximated by, continuous piecewise affine splines, we derive the analytical DGN output space distribution as a function of the product of the DGN's Jacobian singular values raised to a power $\rho$. We dub $\rho$ the {\bf polarity} parameter and prove that $\rho$ focuses the DGN sampling on the modes $(\rho<0)$ or anti-modes $(\rho>0)$ of the DGN output-space probability distribution. 
We demonstrate that nonzero polarity values achieve a better precision-recall (quality-diversity) Pareto frontier than standard methods, such as truncation, for a number of state-of-the-art DGNs.
We also present quantitative and qualitative results on the improvement of overall generation quality (e.g., in terms of the Fr\'echet Inception Distance) for a number of state-of-the-art DGNs, including StyleGAN3, BigGAN-deep, NVAE, for different conditional and unconditional image generation tasks. 
In particular, Polarity Sampling redefines the state-of-the-art for StyleGAN2 on the FFHQ Dataset to FID 2.57, StyleGAN2 on the LSUN Car Dataset to FID 2.27 and StyleGAN3 on the AFHQv2 Dataset to FID 3.95. 
\href{https://bit.ly/polarity-samp}{Colab Demo}\let\thefootnote\relax\footnotetext{$^*$equal contribution}.

\end{abstract}

\begin{figure}[ht]
\begin{center}
\begin{minipage}{\linewidth}
\centering
\begin{minipage}{0.025\linewidth}
\centering
\rotatebox{90}{\footnotesize \hspace{0.5cm}precision (quality)}
\end{minipage}
\begin{minipage}{0.965\linewidth}
\centering
\includegraphics[width=\linewidth]{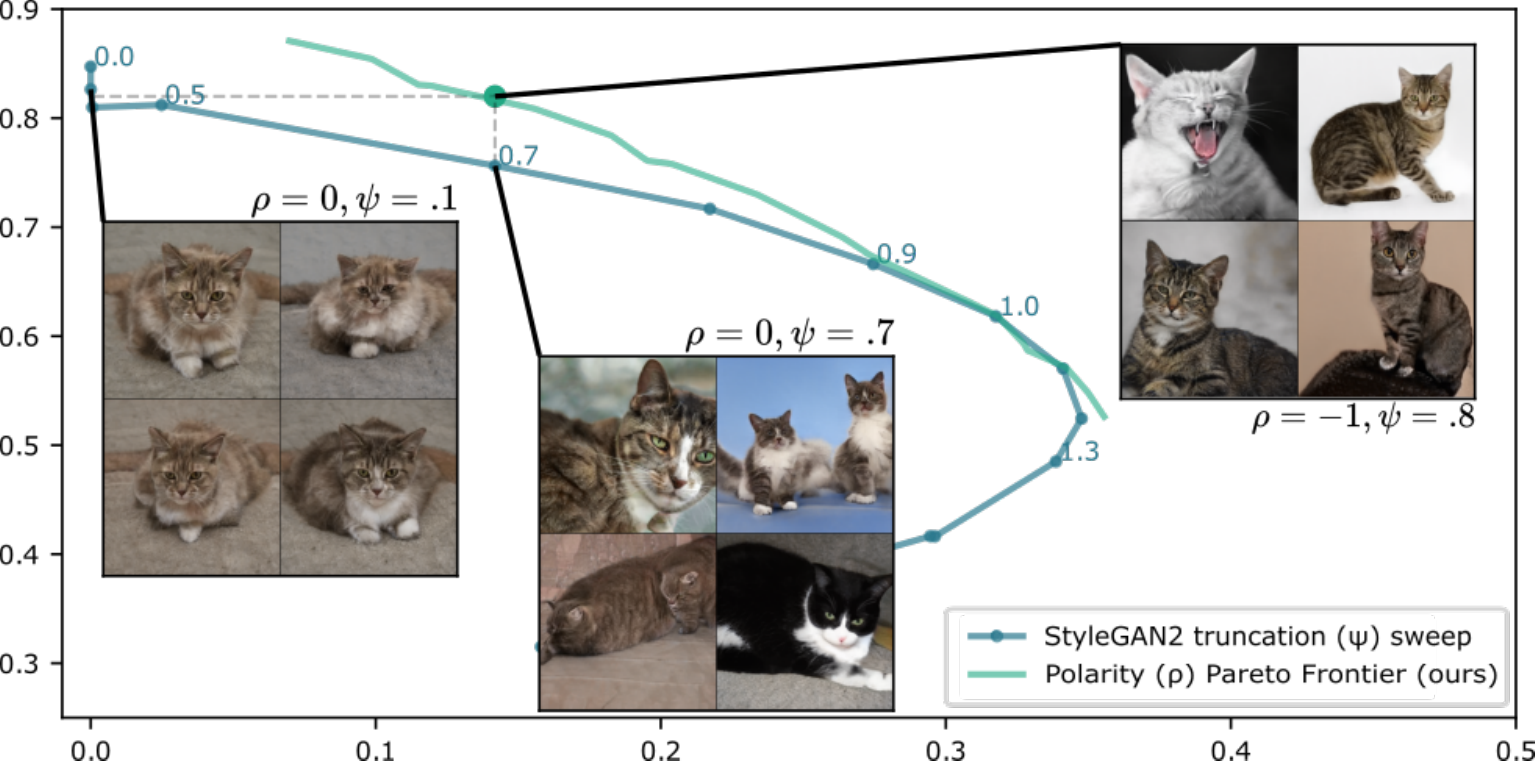}\\[-0.5em]
{\footnotesize recall (diversity)}
\end{minipage}
\end{minipage}
\includegraphics[width=\linewidth]{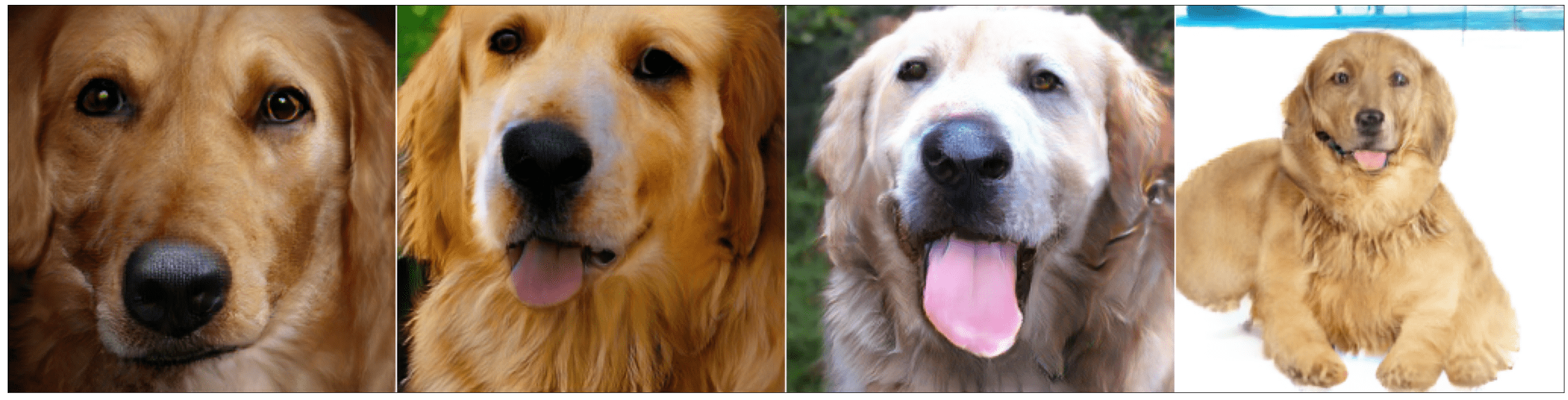}
\includegraphics[width=\linewidth]{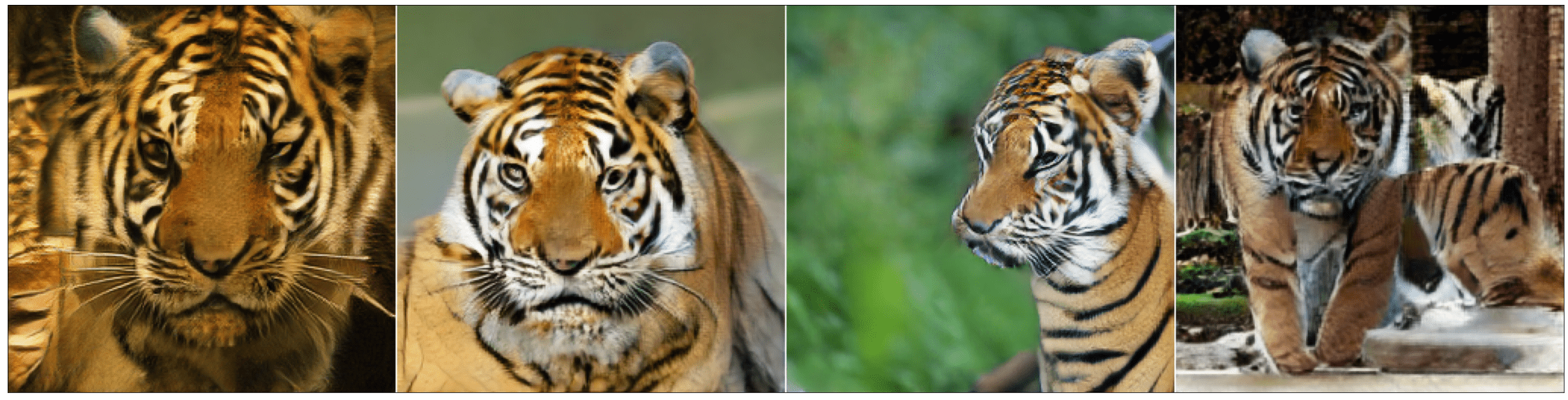}
\includegraphics[width=\linewidth]{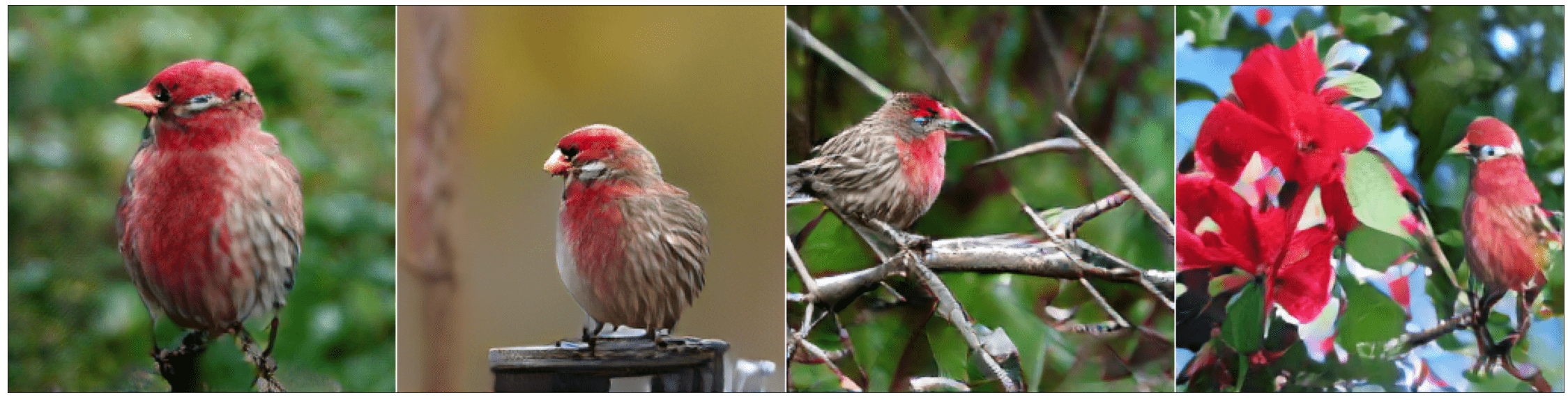}\\
$\rho\ll0$\hspace{1.5cm}$\rho<0$\hspace{1.5cm}$\rho>0$\hspace{1.5cm}$\rho\gg0$
\end{center}
\vspace{-0.5cm}
\caption{
{\bf First row}: Evolution of generation quality and diversity for varying truncation \cite{karras2019style} $\psi$ and polarity $\rho$. 
Polarity Sampling achieves a better Pareto trade-off than truncation, e.g., polarity can be used to achieve a specified recall at higher precision or a specified precision at higher recall, compared to truncation. For additional Pareto examples, see \cref{fig:pareto}.
{\bf Second, Third, and Fourth row:} Samples obtained from BigGAN-deep on \emph{Golden Retriever}, \emph{Tiger} and \emph{House Finch} classes of Imagenet with samples of greater quality $(\rho<0)$ and greater diversity $(\rho>0)$. For examples with LSUN \cite{yu2015lsun}, see \cref{fig:qualitative_alpha_sweep}.
}
\label{fig:teaser}
\vspace{-.5cm}
\end{figure}

\section{Introduction}
\label{sec:intro}

Deep Generative Networks (DGNs) have emerged as the go-to framework for generative modeling of high-dimensional datasets, such as natural images. Within the realm of DGNs, different frameworks can be used to produce an approximation of the data distribution, e.g., Generative Adversarial Networks (GANs) \cite{goodfellow2020generative}, Variational AutoEncoders (VAEs) \cite{kingma2013auto} or flow-based models \cite{prenger2019waveglow}. But despite the different training settings and losses that each of these frameworks aim to minimize, the evaluation metric of choice that is used to characterize the overall quality of generation is the {\em Fr\'echet Inception Distance} (FID) \cite{heusel2017gans}. The FID is obtained by taking the Fr\'echet Distance in the InceptionV3 \cite{szegedy2016rethinking} embedding space between two distributions; the distributions are usually taken to be the training dataset and samples from a DGN trained on the dataset. It has been established in prior work \cite{sajjadi2018assessing} that FID non-linearly combines measures of quality and diversity of the samples, which has inspired further research into disentanglement of these quantities as {\em precision} and {\em recall} \cite{sajjadi2018assessing, kynkaanniemi2019improved} metrics respectively.

Recent state-of-the-art DGNs such as BigGAN \cite{brock2018large}, StyleGAN2/3 \cite{karras2020analyzing,karras2021alias}, and NVAE \cite{vahdat2020nvae}, have reached FIDs nearly as low as one could obtain when comparing subsets of real data with themselves. This has led to the deployment of DGNs in a variety of applications, such as real-world high-quality content generation and data-augmentation. However, it is clear that, {\bf depending on the domain of application, generating samples from the best FID model could be suboptimal}. For example, realistic content generation might benefit more from high-quality (precision) samples, while data-augmentation might benefit more from samples of high-diversity (recall), even if in each case, the overall FID slightly diminishes \cite{giacomello2018doom,islam2021crash}. Therefore, a number of state-of-the-art DGNs have introduced a controllable parameter to trade-off between the precision and recall of the generated samples, e.g., truncated latent space sampling \cite{brock2018large}, interpolating truncation \cite{karras2019style, karras2020analyzing}. However, these methods do not always work ``out-of-the-box'' \cite{brock2018large}, e.g., BigGAN requires orthogonal regularization of the DGN's parameters during training. These methods also lack a clear  theoretical understanding which can limit their deployment for sensitive applications.

{\bf In this paper, we propose a principled solution to control the quality (precision) and diversity (recall) of DGN samples that does not require retraining nor specific conditioning of model training.} 
Our method, termed {\em Polarity Sampling}, builds on our previous work on the analytical form of the learned DGN sample distribution \cite{humayun2021magnet} and introduces a new hyperparameter, that we dub the {\em polarity} $\rho \in \mathbb{R}$, that adapts the latent space distribution for post-training control.
{\bf The polarity parameter provably forces the latent distribution to concentrate on the modes of the DGN distribution, i.e., regions of high probability ($\rho<0$), or on the anti-modes, i.e., regions of low-probability ($\rho>0$); with $\rho=0$ recovering the original DGN distribution.}
The Polarity Sampling process depends only on
the top singular values of the DGN's output Jacobian matrices evaluated at each input sample and can be implemented to perform online sampling. A crucial benefit of Polarity Sampling lies in its theoretical derivation from the analytical DGN data distribution \cite{humayun2021magnet} where the product of the DGN Jacobian matrices singular values -- raised to the power $\rho$ -- provably controls the DGN samples distribution as desired. 
See Fig.~\ref{fig:teaser} for an initial example of Polarity Sampling in action.

Our main contributions are as follows:

\noindent[C1]
We first provide the theoretical derivation of Polarity Sampling based on the singular values of the generator Jacobian matrix. We provide  pseudocode for Polarity Sampling and an approximation scheme to control its computational complexity as desired (\cref{sec:method}).

\noindent[C2]
We demonstrate on a range of DGNs and datasets that Polarity Sampling not only enables one to move on the precision-recall Pareto frontier (\cref{sec:pareto}), i.e., it controls the quality and diversity efficiently, but it also reaches improved FID scores for each model (\cref{sec:sota}).

\noindent[C3]
We leverage the fact that negative Polarity Sampling provides access to the modes of the learned DGN distribution, which enables us to explore several timely and important questions regarding DGNs. We provide visualization of the modes of trained GANs and VAEs (\cref{sec:mode}) and assess the perceptual smoothness around the modes (\cref{sec:PPL}).
\\
\section{Related Work}
\label{sec:related}

\noindent{\bf Deep Generative Networks as Piecewise-Linear Mappings.}~In most DGN settings, once training has been completed, sampling new data points is performed by first sampling latent space samples $\vz_i \in \mathbb{R}^{K}$ from a latent space distribution $\vz_i\sim p_{\vz}$ and then processing those samples throughout a DGN $G:\mathbb{R}^{K} \mapsto \mathbb{R}^{D}$ to obtain the sample $\vx_i\triangleq G(\vz_i),\forall i$. One recent line of research that we will rely on through our study consists in formulating DGNs as Continuous Piecewise Affine (CPA) mappings \cite{montufar2014number,balestriero2018spline}, that be expressed as
\begin{align}
    G(\vz) = \sum_{\omega \in \Omega}(\mA_{\omega}\vz+\vb_{\omega})1_{\{\vz \in \omega\}},\label{eq:CPA}
\end{align}
where $\Omega$ is the input space partition induced by the DGN architecture, $\omega$ is a partition-region where $z$ resides, and $\mA_{\omega}, \vb_{\omega}$ are the corresponding slope and offset parameters. The CPA formulation of \cref{eq:CPA} either represents the exact DGN mapping, when the nonlinearities are CPA e.g. (leaky-)ReLU, max-pooling, or represents a first-order approximation of the DGN mapping. For more background on CPA networks, see \cite{balestriero2020mad}. {\em The key result from \cite{daubi} that we will leverage is that \cref{eq:CPA} is either exact, or can be made close enough to the true mapping $G$, to be considered exact for practical purposes}.

\noindent{\bf Post-Training Improvement of a DGN's Latent Distribution.}
The idea that the training-time latent distribution $p_{\vz}$ might be suboptimal for test-time evaluation has led to multiple research directions to improve the quality of samples post-training.  \cite{tanaka2019discriminator,che2020your} proposed to optimize the samples $\vz \sim p_{\vz}$ based on a Wasserstein discriminator, leading to the {\em Discriminator Optimal Transport} (DOT) method. That is, after sampling a latent vector $\vz$, the latter is repeatedly updated such that the produced datum has greater quality.  \cite{tanielian2020learning} proposes to simply remove the samples that produce data out of the true data manifold. This can be viewed as a binary rejection decision of any new sample $\vz \sim p_{\vz}$. \cite{azadi2018discriminator} were the first to formally introduce rejection sampling based on a discriminator providing a quality estimate used for the rejection sampling of candidate vectors $\vz \sim p_{\vz}$. Replacing rejection sampling with the Metropolis-Hasting algorithm \cite{hastings1970monte} led to the method of
\cite{turner2019metropolis}, coined MH-GAN. An improvement made by \cite{grover2019bias} was to use the {\em Sampling-Importance-Resampling} (SIR) algorithm \cite{rubin1988using}. \cite{issenhuth2021latent} proposes {\em latentRS} which consists in training a WGAN-GP \cite{gulrajani2017improved} on top of any given DGN to learn an improved latent space distribution producing higher-quality samples. \cite{issenhuth2021latent} also proposes {\em latentRS+GA}, where the generated samples from that learned distribution are further improved through gradient ascent.

\noindent{\bf Truncation of the Latent Distribution.}
Latent space truncation was introduced for high-resolution face image generation by \cite{marchesi2017megapixel} as a method of removing generated artifacts. The authors employed a latent prior of $\vz \sim \mathcal{U}[-1,1]$ during training and $\vz \sim \mathcal{U}[-0.5,0.5]$ for qualitative improvement during evaluation. The ``truncation trick'' was formally introduced by \cite{brock2018large} where the authors propose resampling latents $z$ if they exceed a specified threshold for truncation. The authors also use weight orthogonalization during training to make truncation amenable. 
Style-based architectures \cite{karras2019style,karras2020analyzing} introduce a linear interpolation based truncation in the style-space, which is also designed to converge to the average of the dataset \cite{karras2019style}. Ablations for truncation in style-based generators are provided in \cite{kynkaanniemi2019improved}.

\section{Introducing The Polarity Parameter From First Principles}
\label{sec:method}

In this section, we introduce \textit{Polarity Sampling}, a method that enables us to control the generation quality and diversity of DGNs. We will proceed by first expressing the analytical form of DGNs' output distribution (\cref{sec:theory}), and parametrizing the latent space distribution by the singular values of its Jacobian matrix and our {\em polarity} parameter (\cref{sec:polpol}). We provide pseudo-code and an approximation strategy that enables fast sampling (\cref{sec:pseudocode}).

\subsection{Analytical Output-Space Density Distribution}
\label{sec:theory}

Given a DGN $G$, samples are obtained by sampling $G(\vz)$ with a given latent space distribution, as in $\vz \sim p_{\vz}$. This produces samples that will lie on the {\em image} of $G$, the distribution of which is subject to $p_{\vz}$, the DGN latent space partition $\Omega$ and per-region affine parameters $\mA_{\omega},\vb_{\omega}$. We denote the DGN output space distribution as $p_{G}$. Under an injective DGN mapping assumption ($g(z)=g(z') \implies z=z'$) (which holds for various architectures, see, e.g., \cite{puthawala2020globally}) it is possible to obtain the analytical form of the DGN output distribution by $p_{G}$ \cite{humayun2021magnet}. For a reason that will become clear in the next section, we focus here on the case $\vz \sim U(\mathcal{D})$ i.e., using a Uniform latent space distribution over the domain $\mathcal{D}$. Leveraging the Moore-Penrose pseudo inverse \cite{trefethen1997numerical}  $\mA^{\dag}\triangleq (\mA^T\mA)^{-1}\mA^T$, we obtain the following.
\begin{thm}
\label{thm:general_density}
For $\vz \sim U(\mathcal{D})$, the probability density $p_{G}(\vx)$ is given by
\begin{equation}
p_{G}(\vx) \propto \sum_{\omega \in \Omega} \det(\mA_{\omega}^T\mA_{\omega})^{-\frac{1}{2}}\mathds{1}_{\{\mA^{\dag}_{\omega}(\vx-\vb_{\omega}) \in \omega \cap \mathcal{D}\}},
\end{equation}
where $\det$ is the pseudo-determinant, i.e., the product of the nonzero eigenvalues of $\mA_{\omega}^T\mA_{\omega}$. (Proof in Appendix~\ref{proof:general_density}.)
\end{thm}

Note that one can also view $\det(\mA_{\omega}^T\mA_{\omega})^{1/2}$ as the product of the nonzero singular values of $\mA_{\omega}$. Theorem \ref{thm:general_density} is crucial to our development since it demonstrates that the probability of a sample $\vx = g(\vz)$ is proportional to the change in volume $(\det(\mA_{\omega}^T\mA_{\omega})^{1/2})$ produced by the coordinate system $\mA_{\omega}$ of the region $\omega$ in which $\vz$ lies in (recall \cref{eq:CPA}). If 
a region $\omega \in \Omega$ has a slope matrix $\mA_{\omega}$ that contracts the space ($\det(\mA_{\omega}^T\mA_{\omega})<1$) then the output density on that region --- mapped to the output space region $\{\mA_{\omega}\vu+\vb_{\omega}:\vu \in \omega\}$ --- is increased, as opposed to other regions that either do not contract the space as much, or even expand it ($\det(\mA_{\omega}^T\mA_{\omega})>1$). Hence, the concentration of samples in each output space region depends on how that region's slope matrix contracts or expands the space, relative to all other regions.

\subsection{Controlling the Density Concentration with a Single Parameter}
\label{sec:polpol}

From \cref{thm:general_density} we can  directly obtain an explicit parametrization of $p_{\vz}$ that enables us to control the distribution of samples in the output space, i.e., to control $p_{G}$. In fact, note that one can sample from the mode of the DGN distribution by employing $\vz \sim U(\omega^*), \omega^* = \argmin_{\omega \in \Omega}\det(\mA_{\omega}^T\mA_{\omega})$. Alternatively, one can sample from the region of lowest probability, i.e., the anti-mode, by employing $\vz \sim U(\omega^*), \omega^* = \argmax_{\omega \in \Omega}\det(\mA_{\omega}^T\mA_{\omega})$.
This directly leads to our Polarity Sampling method that adapts the latent space distribution based on the per-region pseudo-determinants.

\begin{cor}
\label{cor:pol_density}
The latent space distribution
\begin{equation}
    p_{\rho}(\vz) \propto \sum_{\omega \in \Omega}\det(\mA_{\omega}^T\mA_{\omega})^{\frac{\rho}{2}} \mathds{1}_{\{\vz \in \omega\}},\label{eq:latent_distribution}
\end{equation}
where ${\rho}\in\mathbb{R}$ is the polarity parameter, produces the DGN output distribution
\begin{equation}
p_{G}(\vx)\propto \sum_{\omega \in \Omega} \det(\mA_{\omega}^T\mA_{\omega})^{\frac{{\rho}-1}{2}}\mathds{1}_{\{\mA^{\dag}_{\omega}(\vx-\vb_{\omega}) \in \omega \cap \mathcal{D}\}},\label{eq:density_rho}
\end{equation}
which falls back to the standard DGN distribution for ${\rho}=0$, to sampling of the mode(s)
for ${\rho} \to -\infty$ and to sampling of the anti-mode(s) for ${\rho} \to \infty$. (Proof in Appendix~\ref{proof:pol_density}.)
\end{cor}
%
%
%
Polarity Sampling consists of using the latent space distribution \cref{eq:latent_distribution} with a polarity parameter $\rho$, that is either negative, concentrating the samples toward the mode(s) of the DGN distribution $p_{G}$, positive, concentrating the samples towards the anti-modes(s) of the DGN distribution $p_{G}$ or zero, which removes the effect of polarity.
Note that Polarity Sampling changes the output density in a continuous fashion. Its practical effect, as we will see in Sec.~\ref{sec:pareto}, is to control the quality and diversity of the obtained samples.

\subsection{Approximation and Implementation}
\label{sec:pseudocode}

We now provide the details and pseudocode for the Polarity Sampling procedure that implements \cref{cor:pol_density}.

\noindent{\bf Computing the $\mA_{\omega}$ Matrix.}~The per-region slope matrix as in \cref{eq:CPA}, can be obtained given any DGN by first sampling a latent vector $\vz \in \omega$, and then obtaining the Jacobian matrix of the DGN $\mA_{\omega} = \mJ_{\vz}G(\vz),\forall \vz \in \omega$. This has the benefit of directly employing automatic differentiation libraries and thus does not require any exhaustive implementation nor derivation. Computing $\mJ_{\vz}G(\vz)$ of a generator is not uncommon in practice, e.g., it is employed during path length regularization of StyleGAN2 \cite{karras2020analyzing}.

\noindent{\bf Discovering the Regions $\omega \in \Omega$.}~As per \cref{eq:latent_distribution}, we need to obtain the singular values of $\mA_{\omega}$ (see next paragraph) for each region $\omega \in \Omega$. This is often a complicated task, especially for state-of-the-art DGNs that can have a partition $\Omega$ whose number of regions grows with the architecture depth and width \cite{montufar2021sharp}.
Furthermore, checking if $\vz \in \omega$ requires one to solve a linear program \cite{fischetti2017deep}, which is expensive. As a result, we develop an approximation that consists of sampling many $\vz \sim U(\mathcal{D})$ vectors from the latent space (hence our uniform prior assumption in \cref{cor:pol_density}), and computing their corresponding matrices $\mA_{\omega(\vz)}$. This way, we are guaranteed that $\mA_{\omega(\vz)}$ corresponds to the slope of the region $\omega$ in which $\vz$ falls in, removing the need to check whether $\vz \in \omega$. We do so over $N$ samples obtained uniformly from the DGN latent space (based on the original latent space domain). Selection of $N$ can impact performance as this exploration needs to discover as many regions from $\Omega$ as possible.

\noindent{\bf Singular Value Computation.}~Computing the singular values of $\mA_{\omega}$ is an $\mathcal{O}(\min(K,D)^3)$ operation \cite{golub1971singular}. However, not all singular values might be relevant, e.g., the smallest singular values that are nearly constant across regions $\omega$ can be omitted without altering \cref{cor:pol_density}. Hence, we employ only the top-$k$ singular values of $\mA_{\omega}$ to speed up singular value computation to $\mathcal{O}(Dk^2)$, details provided in \cref{sec:times}. (Further approximation could be employed if needed, e.g., power iteration \cite{packard1988power}).

\begin{figure}[t!]
\begin{center}
\begin{minipage}{0.02\linewidth}
\centering
\rotatebox{90}{\footnotesize \hspace{0.7cm}FID}
\end{minipage}
\begin{minipage}{0.48\linewidth}
\centering
\includegraphics[width=\linewidth]{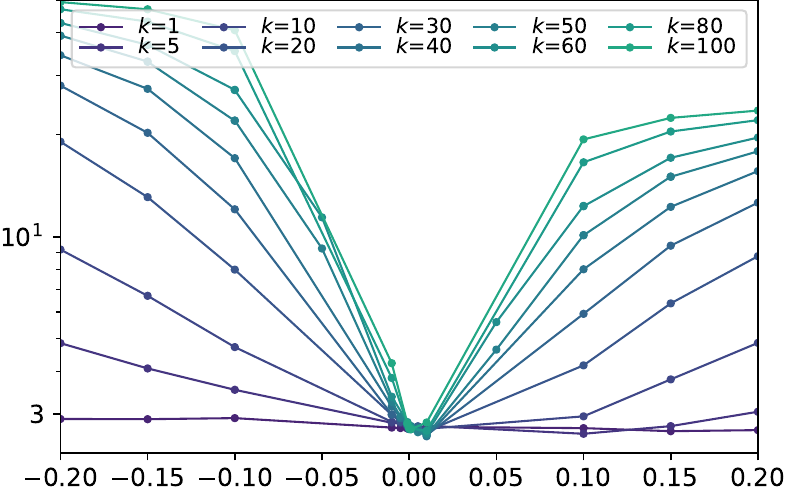}\\[-0.5em]
{\footnotesize polarity $(\rho)$}
\end{minipage}
\begin{minipage}{0.48\linewidth}
\centering
\includegraphics[width=\linewidth]{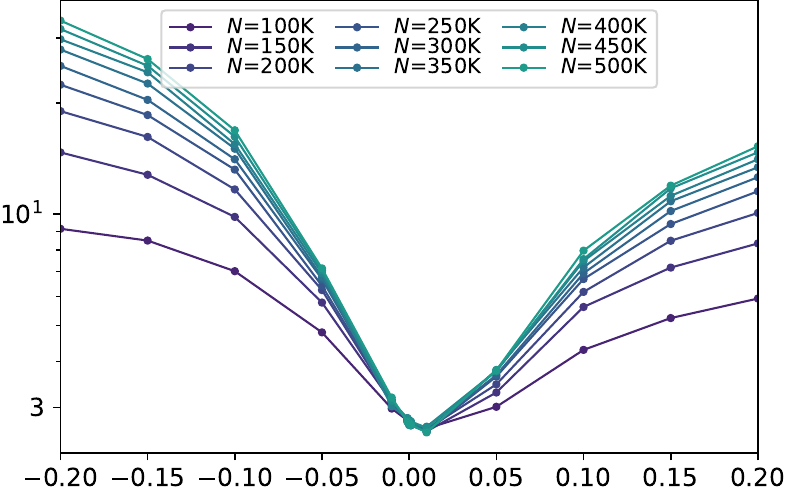}\\[-0.5em]
{\footnotesize polarity $(\rho)$}
\end{minipage}
\end{center}
\vspace{-0.5cm}
\caption{ Effect of Polarity Sampling on FID of a StyleGAN2-F model pretrained on FFHQ for varying number of top-$k$ singular values ({\bf left}) and varying number of latent space samples $N$ used to obtain per-region slope matrix $\mA_{\omega}$ singular values ({\bf right}) (recall \cref{sec:pseudocode,alg:sampling}). The trend in FIDs to evaluate the impact of $\rho$ stabilizes when using around $k=40$ singular values and $N\approx$200,000 latent space samples. For the effect of $k$ and $N$ on precision and recall, see \cref{fig:topk_prec_recall}.
}
\label{fig:topk_dets}
\end{figure}

While the required number of latent space samples $N$ and the number of top singular values $k$ might seem to be a limitation of Polarity Sampling, we have found in practice that $N$ and $k$ for state-of-the-art DGNs can be set at $N \approx 200K$, $k \in [30,100]$. We conduct a careful ablation study and demonstrate the impact of different choices for $N$ and $k$ in \cref{fig:topk_dets,tab:ablation_N,tab:ablation_topk} in \cref{sec:effect_N_k}. Computation times and software/hardware details are provided in \cref{sec:times}. To reduce round-off errors that can occur for extreme values of $\rho$, we compute the product of singular values in log-space, as shown in \cref{alg:sampling}. 

\begin{algorithm}[t!]
\caption{
Polarity Sampling procedure with polarity $\rho$; online version and 2D examples in Appendix. \cref{alg:online} and \cref{2d_examples}. For implementation details, see \cref{sec:pseudocode}.}\label{alg:sampling}
\begin{algorithmic}
\Require $K>0,S>0,N \gg S,G,\mathcal{D},\rho \in \mathbb{R}$
\State $\gZ,\gS,\gR \gets [],[],[]$
\For{$n=1,\dots,N$}
    \State $\vz \sim U(\mathcal{D})$
    \State $\sigma = {\rm SingularValues}(\mJ_{\vz}G(\vz),{\rm decreasing=True})$
    \State $\gZ.{\rm append}(\vz)$
    \State $\gS.{\rm append}(\rho \sum_{k=1}^{K}\log(\sigma[k]+\epsilon))$
\EndFor
\For{$n=1,\dots,S$}
    \State $i \sim {\rm Categorical}({\rm prob}={\rm softmax}(\gS))$
    \State $\gR.{\rm append}(\gZ[i])$
\EndFor
\Ensure $\gR$
\end{algorithmic}
\end{algorithm}

We summarize how to obtain $S$ samples using the above steps in the pseudocode given in \cref{alg:sampling} and provide an efficient solution to reduce the memory requirement incurred when computing the large matrix $\mA_{\omega}$ in \cref{sec:memory}. We also provide an implementation that enables online sampling in \cref{alg:online} (\cref{sec:online_algorithm}). It is also possible to control the DGN prior $p_{\vz}$ with respect to a different space than the data-space e.g. inception-space, or with a different input space than the latent-space e.g. style-space in StyleGAN2/3. This incurs no changes in \cref{alg:sampling} except that the DGN is now considered to be either a subset of the original one, or to be composed with a VGG/InceptionV3 network. We provide the implementation details for style-space, VGG-space, and Inception-space in \cref{sec:other_space}. In those cases, the partition $\Omega$ and the per-region mapping parameters $\mA_{\omega},\vb_{\omega}$ are the ones of the corresponding sub-network or composition of networks (recall \cref{eq:CPA}). Polarity Sampling adapts the DGN prior distribution to obtain the modes or anti-modes with respect to the considered output spaces.

\section{Controlling Precision, Recall, and FID via Polarity}
\label{sec:experiments}

We now provide empirical validation of Polarity Sampling with an extensive array of experiments.
Since calculation of distribution metrics such as FID, precision, and recall are sensitive to image processing nuances, we use each model's original code repository except for BigGAN-deep on ImageNet \cite{deng2009imagenet}, for which we use the evaluation pipeline specified for
ADM \cite{dhariwal2021diffusion}. 
For NVAE (trained on colored-MNIST \cite{arjovsky2019invariant}), we use a modified version of the StyleGAN3 evaluation pipeline. Precision and recall metrics are all based on the implementation of \cite{kynkaanniemi2019improved}. Metrics in \cref{tab:main_result} are calculated for 50K training samples to be able to compare with existing latent reweighing methods. For all other results, the metrics are calculated using $\min\{N_{D},100K\}$ training samples, where $N_{D}$ is the number of samples in the dataset. 


\subsection{Polarity Efficiently Parametrizes the Precision-Recall Pareto Frontier}
\label{sec:pareto}

\begin{figure}[t!]
\begin{center}
\begin{minipage}{0.02\linewidth}
\centering
\rotatebox{90}{\small
\hspace{1cm}Precision\hspace{2.7cm}Precision\hspace{2.7cm}Precision}
\end{minipage}
\begin{minipage}{0.48\linewidth}
\centering
\includegraphics[width=\linewidth]{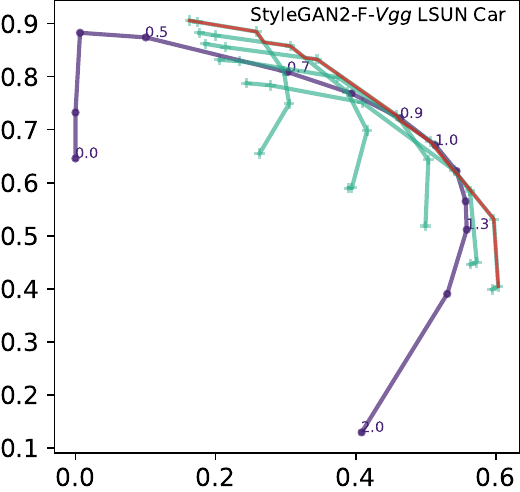}\\
\includegraphics[width=\linewidth]{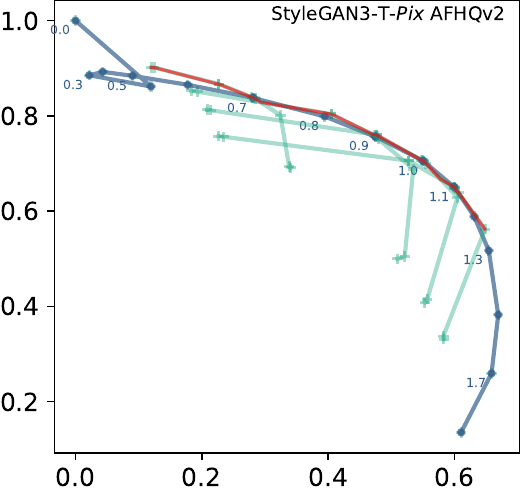}\\
\includegraphics[width=\linewidth]{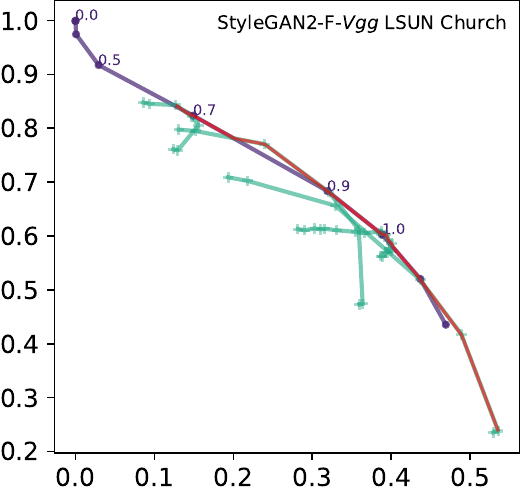}\\[-0.4em]
{\small Recall}
\end{minipage}
\begin{minipage}{0.48\linewidth}
\centering
\includegraphics[width=\linewidth]{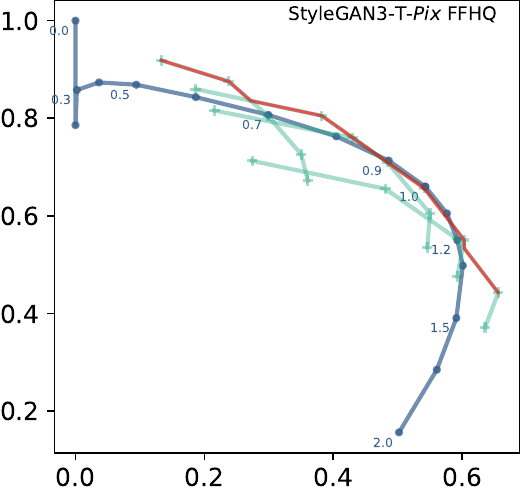}\\
\includegraphics[width=\linewidth]{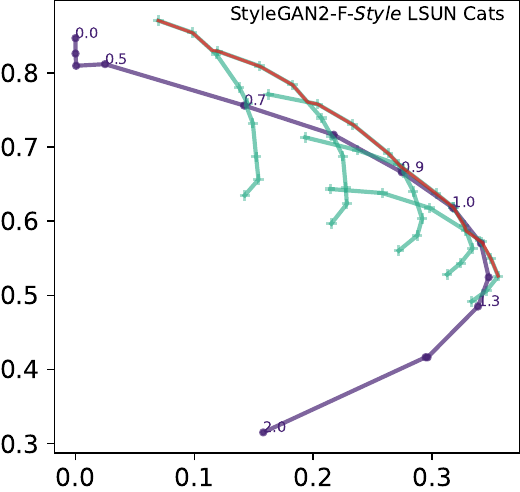}\\
\includegraphics[width=\linewidth]{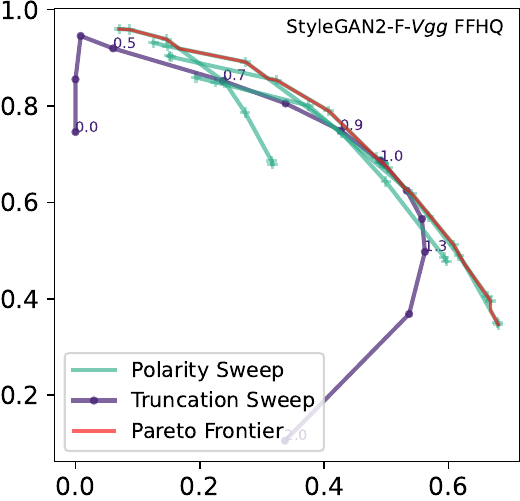}\\[-0.4em]
{\small Recall}
\end{minipage}
\end{center}
\vspace{-0.5cm}
\caption{Pareto frontier of the precision-recall metrics can be obtained solely by varying the polarity parameter, for any given truncation level. We depict here six different models and datasets. Results for additional models and datasets are provided in \cref{fig:teaser} and \cref{fig:more_pareto}. 
}
\label{fig:pareto}
\end{figure}

\begin{figure*}[t!]
\centering
\begin{minipage}{0.02\linewidth}
      \hfill
\end{minipage}
\begin{minipage}{0.10\linewidth}
\centering
{\scriptsize (modes\reflectbox{$\to$})}\\
      $\rho=-2$
\end{minipage}
\begin{minipage}{0.10\linewidth}
\centering\vspace{0.28cm}
      $\rho=-1$
\end{minipage}
\begin{minipage}{0.10\linewidth}
\centering\vspace{0.28cm}
      $\rho=-0.5$
\end{minipage}
\begin{minipage}{0.10\linewidth}
\centering\vspace{0.28cm}
      $\rho=-0.2$
\end{minipage}
\begin{minipage}{0.10\linewidth}
\centering
      {\scriptsize (baseline)}\\
      $\rho=0$
\end{minipage}
\begin{minipage}{0.10\linewidth}
\centering\vspace{0.28cm}
      $\rho=0.2$
\end{minipage}
\begin{minipage}{0.10\linewidth}
\centering\vspace{0.28cm}
      $\rho=0.5$
\end{minipage}
\begin{minipage}{0.10\linewidth}
\centering\vspace{0.28cm}
      $\rho=1$
\end{minipage}
\begin{minipage}{0.10\linewidth}
\centering
      {\scriptsize ($\to$ anti-modes)}\\
      $\rho=2$
\end{minipage}\\
\begin{minipage}{0.02\linewidth}
      \rotatebox[origin=c]{90}{LSUN Cars} 
\end{minipage}
\begin{minipage}{0.97\textwidth}
      \includegraphics[width=\linewidth]{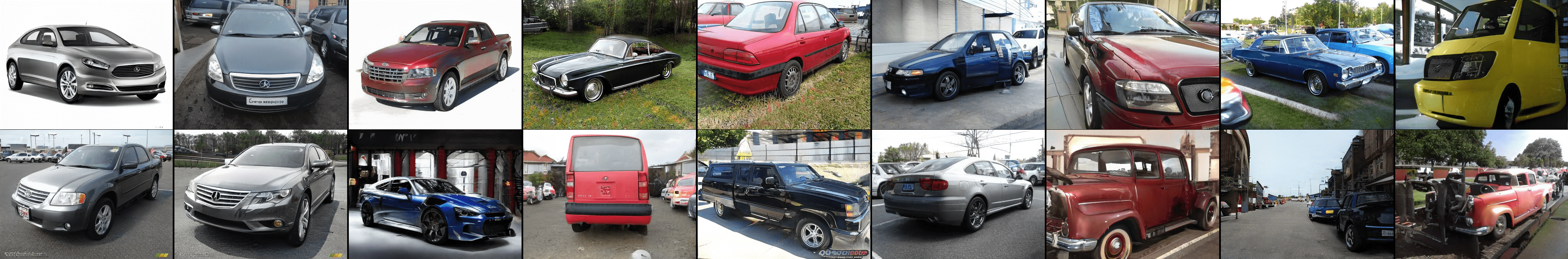}
\end{minipage}\\
\begin{minipage}{0.02\linewidth}
      \rotatebox[origin=c]{90}{LSUN Cats} 
\end{minipage}
\begin{minipage}{0.97\textwidth}
      \includegraphics[width=\linewidth]{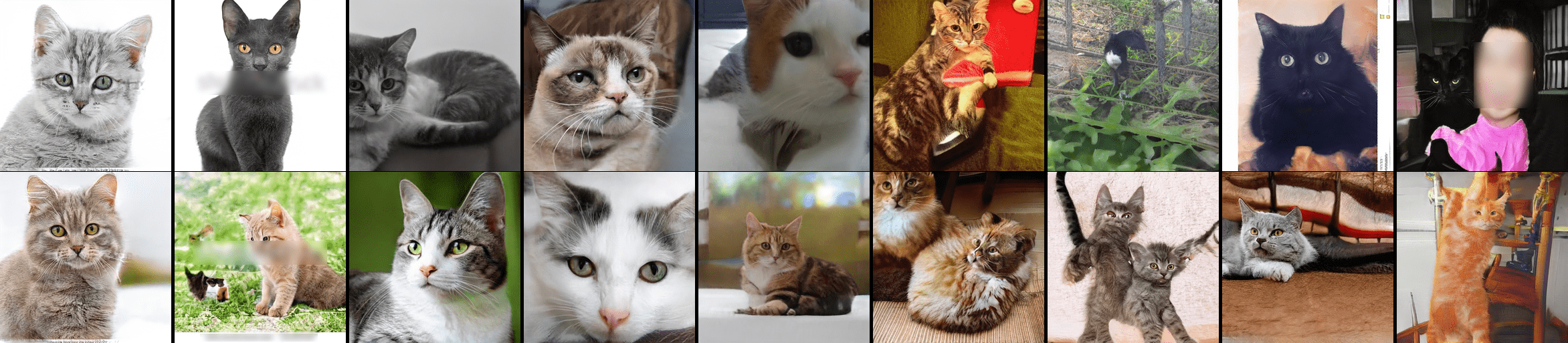}
\end{minipage}\\
\begin{minipage}{0.02\linewidth}
      \rotatebox[origin=c]{90}{LSUN Church} 
\end{minipage}
\begin{minipage}{0.97\textwidth}
      \includegraphics[width=\linewidth]{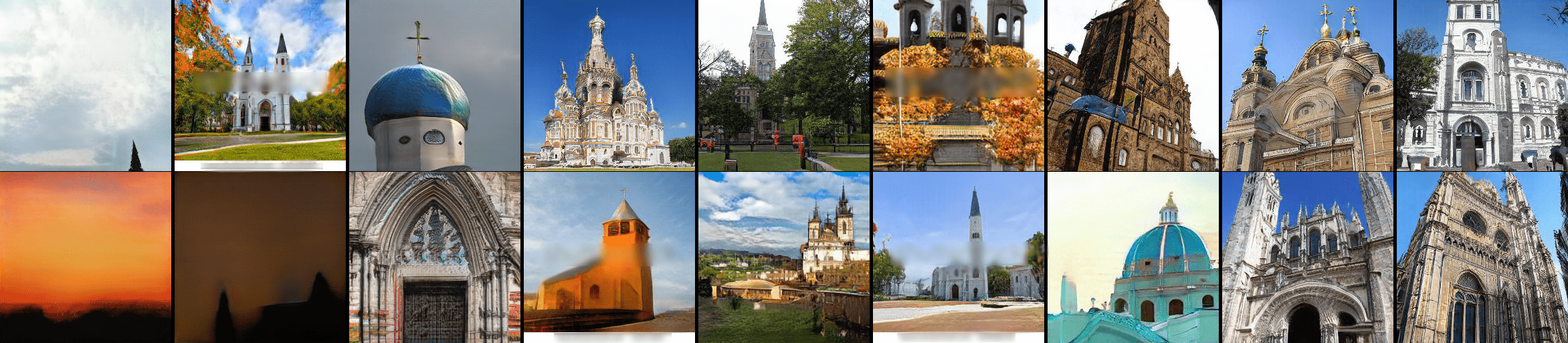}
\end{minipage}
\vspace{-0.2cm}
\caption{Curated samples of cars and cats for Polarity Sampling in style-space, and church for Polarity Sampling in pixel-space.
(Qualitative comparison with truncation sweep in \cref{fig:truncation_sweep_images} and nearest training samples in \cref{fig:nearest} in the Appendix.) None of the images correspond to training samples, as we discuss in \cref{sec:mode}.
}
\label{fig:qualitative_alpha_sweep}
\end{figure*}

As we have discussed above, Polarity Sampling can explicitly sample from the modes or anti-modes of any learned DGN distribution. Since the DGN is trained to fit the training distribution, sampling from the modes and anti-modes correspond to sampling from regions of the data manifold that are approximated better/worse by the DGN 
. Therefore, Polarity Sampling is an efficient parameterization of the trade-off between precision and recall of generation \cite{kynkaanniemi2019improved} since regions with higher precision are regions where the manifold approximation is more accurate. 

As experimental proof, we provide in \cref{fig:pareto} the precision-recall trade-off when sweeping polarity, and compare it with truncation \cite{karras2019style} for pretrained StyleGAN\{2,3\} architectures. We see that Polarity Sampling offers a competitive alternative to truncation for controlling the precision-recall trade-off of DGNs across datasets and models.
For any given precision, the $\rho$ parameter allows us to reach greater recall than what is possible via latent space truncation \cite{karras2019style}. And conversely, for any given recall, it is possible to reach a higher precision than what can be attained using latent space truncation. We see that diversity collapses rapidly for latent truncation compared to Polarity Sampling, across all architectures, which is a major limitation. In addition to that, controlling both truncation and polarity allows us to further extend the Pareto frontier for all of our experiments.

Apart from the results presented here, we also see that polarity can be used to effectively control the precision-recall trade-off for BigGAN-deep \cite{brock2018large} and ProGAN \cite{karras2017progressive}. ProGAN unlike BigGAN and StyleGAN, is not compatible with truncation based methods, i.e., latent space truncation has negligible effect on precision-recall. Hence, polarity offers a great benefit over those existing solutions: Polarity Sampling can be applied regardless of training or controllability factors that are preset in the DGN design. We provide additional results in \cref{appendix:extra}.


\subsection{Polarity Improves Any DGN's FID}
\label{sec:sota}

We saw in \cref{sec:pareto} that polarity can be used to control quality versus diversity in a meaningful and controllable manner. In this section, we connect the effect of polarity with FID. Recall that the FID metric nonlinearly combines quality and diversity \cite{sajjadi2018assessing} into a distribution distance measure. Since polarity allows us to control the output distribution of the DGN, an indirect result of polarity is the reduction of FID by matching the inception embedding distribution of the DGN with that of the training set distribution. Recall that $\rho=0$ recovers the baseline DGN sampling; for all the state-of-the-art methods in question, we reach lower (better) FID by using a nonzero polarity. In \cref{tab:prior_art}, we compare Polarity Sampling with state-of-the-art solutions that propose to improve FID by learning novel DGN latent space distributions, as were discussed in \cref{sec:related}. We see that for a StyleGAN2 pre-trained on the LSUN church \cite{yu2015lsun} dataset, by increasing the diversity ($\rho=0.2$) of the VGG embedding distribution, Polarity Sampling surpasses the FID of methods reported in literature that post-hoc improves quality of generation.

\begin{table}[t!]
\centering
\def\arraystretch{0.5}%
\begin{tabular}{@{}lrrr@{}}
\toprule
\multicolumn{4}{r}{\textbf{LSUN Church 256$\times$256}} \\ \midrule  \hline
StyleGAN2 variant & \multicolumn{1}{l}{FID $\downarrow$} & \multicolumn{1}{l}{Prec $\uparrow$} & \multicolumn{1}{l}{Recall $\uparrow$} \\ \midrule
Standard & 6.29 & .60 & .51 \\
SIR$^\dagger$ \cite{rubin1988using} & 7.36 & .61 & \textbf{.58} \\
DOT$^\dagger$ \cite{tanaka2019discriminator} & 6.85 & .67 & .48 \\
latentRS$^\dagger$ \cite{issenhuth2021latent} & 6.31 & .63 & .58 \\
latentRS+GA$^\dagger$ \cite{issenhuth2021latent} & 6.27 & \textbf{.73} & .43 \\
$\rho$-sampling 0.2  & \textbf{6.02} & .57 & .53 \\ \bottomrule
\end{tabular}
\vspace{-0.2cm}
\caption{
Comparison of 
Polarity Sampling with latent reweighting techniques from literature. FID, Precision and Recall is calculated using 50,000 samples. $^\dagger$Metrics reported from papers due to unavailability of code. $^\dagger$Precision-recall is calculated with $1024$ samples only.
}
\label{tab:prior_art}
\end{table}

\begin{table*}[t!]
\centering
\setlength\tabcolsep{0.5em}
\begin{tabular}{llrrrllrr}
\hline
Model & FID $\downarrow$ & \multicolumn{1}{l}{Precision $\uparrow$} & \multicolumn{1}{l}{Recall $\uparrow$} &  & Model & FID $\downarrow$ & \multicolumn{1}{l}{Precision $\uparrow$} & \multicolumn{1}{l}{Recall $\uparrow$} \\ \hline
\multicolumn{4}{r}{\textbf{LSUN Church 256$\times$256}} &  & \multicolumn{4}{r}{\textbf{LSUN Cat 256$\times$256}} \\ \cline{1-4} \cline{6-9} 
DDPM$^\dagger$ \cite{ho2020denoising} & 7.86 & - & - &  & ADM (dropout)$^\dagger$ & {\bf 5.57} & 0.63 & \textbf{0.52} \\
StyleGAN2 & 3.97 & 0.59 & 0.39 &  & StyleGAN2 & 6.49 & 0.62 & 0.32 \\
+ $\rho$-sampling Vgg 0.001 & 3.94 & 0.59 & 0.39 &  & + $\rho$-sampling Pix 0.01 & 6.44 & 0.62 & 0.32 \\
+ $\rho$-sampling Pix -0.001 & \textbf{3.92} & \textbf{0.61} & \textbf{0.39} &  & + $\rho$-sampling Sty -0.1 & 6.39 & \textbf{0.64} & 0.32 \\ \cline{1-4} \cline{6-9} 
\multicolumn{4}{r}{\textbf{LSUN Car 512$\times$384}} &  & \multicolumn{4}{r}{\textbf{FFHQ 1024$\times$1024}} \\ \cline{1-4} \cline{6-9} 
StyleGAN$^\dagger$ & 3.27 & \textbf{0.70} & 0.44 &  & StyleGAN2-E & 3.31 & \textbf{0.71} & 0.45 \\
StyleGAN2 & 2.34 & 0.67 & 0.51 &  & Projected GAN$^\dagger$ \cite{sauer2021projected} & 3.08 & 0.65 & 0.46 \\
+ $\rho$-sampling Vgg -0.001 & 2.33 & 0.68 & 0.51 &  & StyleGAN3-T & 2.88 & 0.65 & 0.53 \\
+ $\rho$-sampling Sty 0.01 & \textbf{2.27} & 0.68 & \textbf{0.51} &  & + $\rho$-sampling Vgg -0.01 & 2.71 & 0.66 & \textbf{0.54} \\
+ $\rho$-sampling Pix 0.01 & 2.31 & 0.68 & 0.50 &  &  &  & &  \\
\cline{1-4}
\multicolumn{4}{r}{\textbf{ImageNet 256$\times$256}} &  & StyleGAN2-F & 2.74 & 0.68 & 0.49 \\ \cline{1-4}
DCTransformer$^\dagger$ \cite{nash2021generating} & 36.51 & 0.36 & 0.67 &  & + $\rho$-sampling Ic3 0.01 & \textbf{2.57} & 0.67 & 0.5 \\
VQ-VAE-2$^\dagger$ \cite{razavi2019generating} & 31.11 & 0.36 & 0.57 &  & + $\rho$-sampling Pix 0.01 & 2.66 & 0.67 & 0.5 \\
SR3 $^\dagger$\cite{saharia2021image} & 11.30 & - & - &  &  &  &  &  \\ \cline{6-9} 
IDDPM$^\dagger$\cite{nichol2021improved} & 12.26 & 0.70 & 0.62 &  & \multicolumn{4}{r}{\textbf{AFHQv2 512$\times$512}} \\ \cline{6-9} 
ADM$^\dagger$\cite{dhariwal2021diffusion} & 10.94 & 0.69 & \textbf{0.63} &  & StyleGAN2$^\dagger$ & 4.62 & - & - \\
ICGAN+DA$^\dagger$\cite{casanova2021instanceconditioned} & 7.50 & - & - &  & StyleGAN3-R$^\dagger$ & 4.40 & - & - \\
BigGAN-deep & 6.86 & 0.85 & 0.29 &  & StyleGAN3-T & 4.05 & 0.70 & 0.55 \\
+  $\rho$-sampling Pix 0.0065 & 6.82 & \textbf{0.86} & 0.29 &  & + $\rho$-sampling Vgg -0.001 & \textbf{3.95} & \textbf{0.71} & \textbf{0.55} \\
ADM+classifier guidance & \textbf{4.59} & 0.82 & 0.52 &  &  &  &  &  \\ \hline
\end{tabular}

\vspace{-0.2cm}
\caption{$^\dagger$Paper reported metrics. We observe that moving away from $\rho=0$, Polarity Sampling improves FID across models and datasets, empirically validating that the top singular values of a DGN's Jacobian matrices contain meaningful information to improve the overall quality of generation}.
\label{tab:main_result}
\end{table*}

In \cref{tab:main_result}, we present for LSUN \{Church, Car, Cat\} \cite{yu2015lsun}, ImageNet \cite{deng2009imagenet}, FFHQ \cite{karras2019style}, and AFHQv2 \cite{choi2020stargan,karras2021alias} improved FID obtained \textit{solely by changing the polarity $\rho$} of a state-of-the-art DGN. This implies that Polarity Sampling
provides an efficient solution to adapt the DGN latent space.

We observe that, given any specific setting, $\rho \not = 0$ always improves a model's FID. We see that in a case specific manner, both positive and negative $\rho$ improves the FID.For StyleGAN2-F trained on FFHQ, \textit{increasing the diversity} of the inception space embedding distribution helps reach a new state-of-the-art FID. By \textit{increasing the precision} of StyleGAN3-T via Polarity Sampling in the Vgg space, we are able to surpass the FID of baseline StyleGAN2-F \cite{karras2021alias}. We observe that controlling the polarity of the InceptionV3 embedding distribution of StyleGAN2-F gives the most significant gains in terms of FID. This is due to the fact that the Frechet distance between real and generated distributions is directly affected while performing Polarity Sampling in the Inception space. We provide generated samples in \cref{fig:qualitative_alpha_sweep} varying the style-space $\rho$ for LSUN cars and LSUN cats, whereas varying the pixel-space $\rho$ for LSUN Church. It is clear that $\rho<0$ i.e. sampling closer to the DGN distribution modes produce samples of high visual quality, while $\rho>0$ i.e. sampling closer to the regions of low-probability produce samples of high-diversity, with some samples which are off the data manifold due to the approximation quality of the DGN in that region. Using Polarity Sampling, we are able to advance the state-of-the-art performance on three different settings: for StyleGAN2 on the FFHQ \cite{karras2019style} Dataset to FID 2.57, StyleGAN2 on the LSUN \cite{yu2015lsun} Car Dataset to FID 2.27, and StyleGAN3 on the AFHQv2 \cite{karras2021alias} Dataset to FID 3.95. For additional experiments with ProGAN, and NVAE under controlled training and reference dataset distribution shift, see \cref{appendix:extra}.

\vspace{-0.2cm}
\section{New Insights into DGN Distributions}

\vspace{-0.2cm}
In \cref{sec:experiments} we demonstrated that Polarity Sampling 
is a practical method to manipulate DGN output distributions to control their quality and diversity. 
We now demonstrate that Polarity Sampling has more foundational theoretical applications as well.
In particular, we dive into several timely questions regarding DGNs that can be probed using our framework. 
\vspace{-0.2cm}
\subsection{Are GAN/VAE Modes Training Samples?}
\label{sec:mode}
\begin{figure}[t!]
\begin{center}
\begin{minipage}{0.32\linewidth}
\begin{minipage}{\columnwidth} \centering {\tiny BigGAN Samoyed}  \end{minipage}
\includegraphics[width=.99\columnwidth]{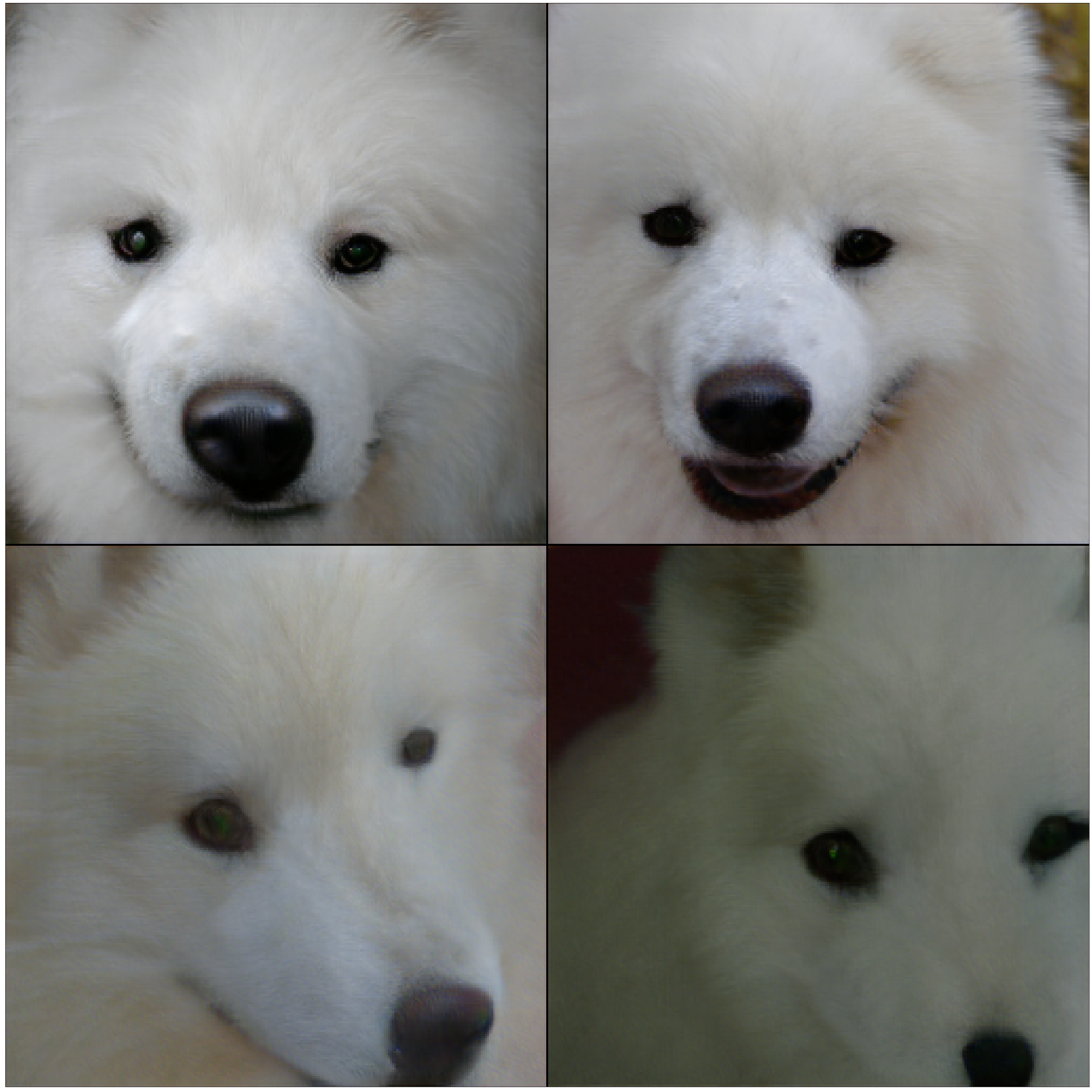}\\
\includegraphics[width=.99\columnwidth]{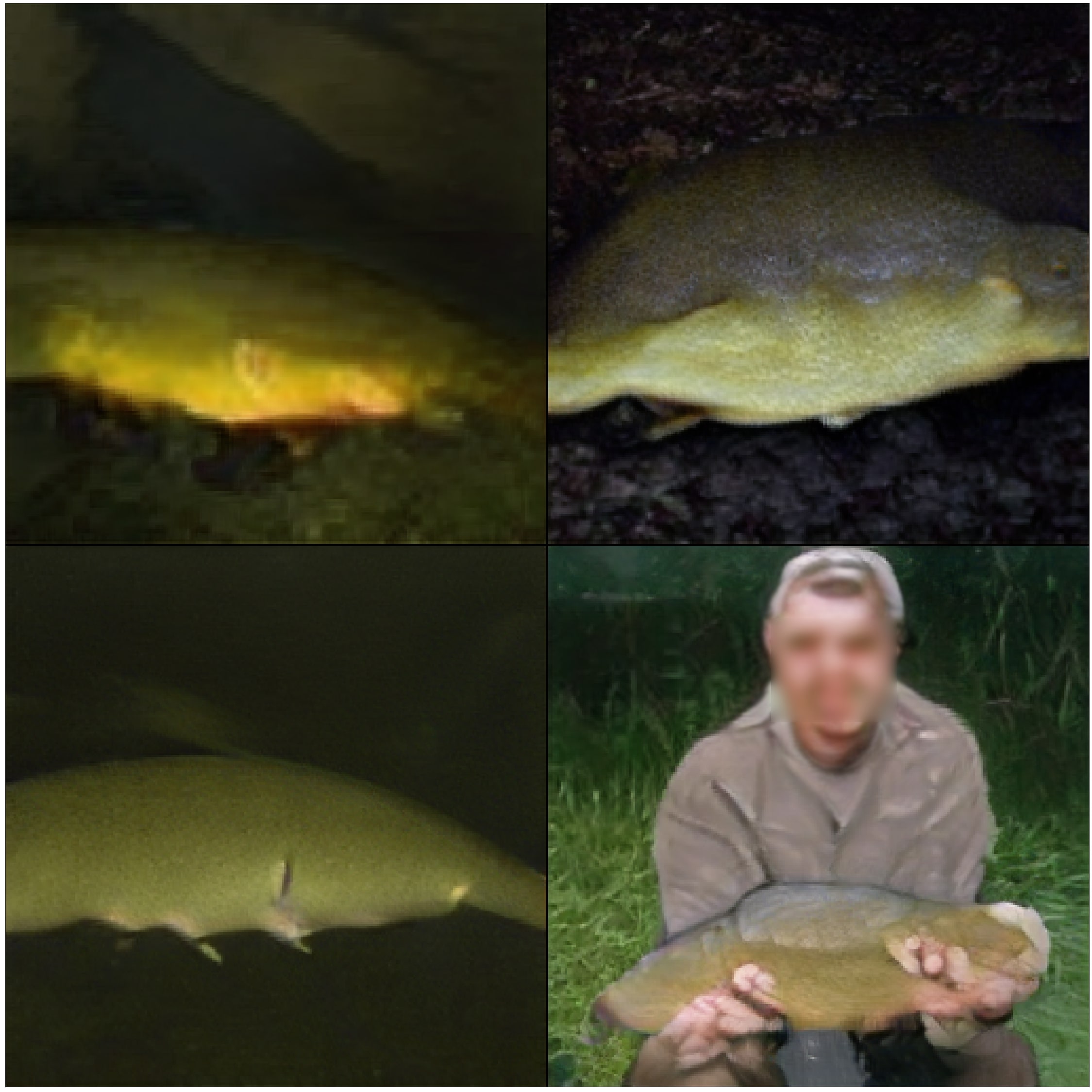}
\begin{minipage}{\columnwidth} \centering \vspace{-1em} {\tiny BigGAN Tench}  \end{minipage}
\end{minipage}
\begin{minipage}{0.32\linewidth}
\begin{minipage}{\columnwidth} \centering {\tiny BigGAN Flamingo}  \end{minipage}
\includegraphics[width=.99\columnwidth]{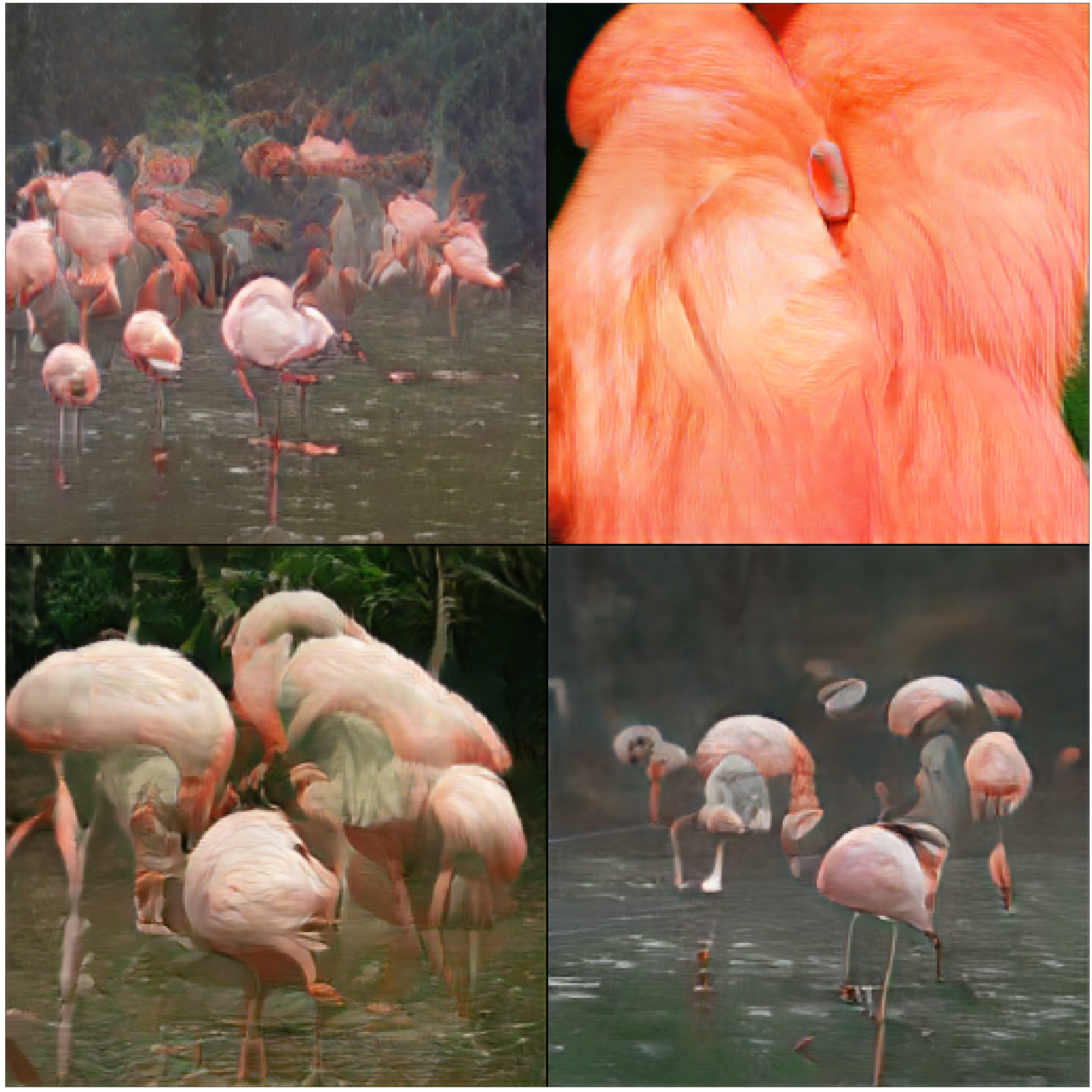}
\includegraphics[width=.99\columnwidth]{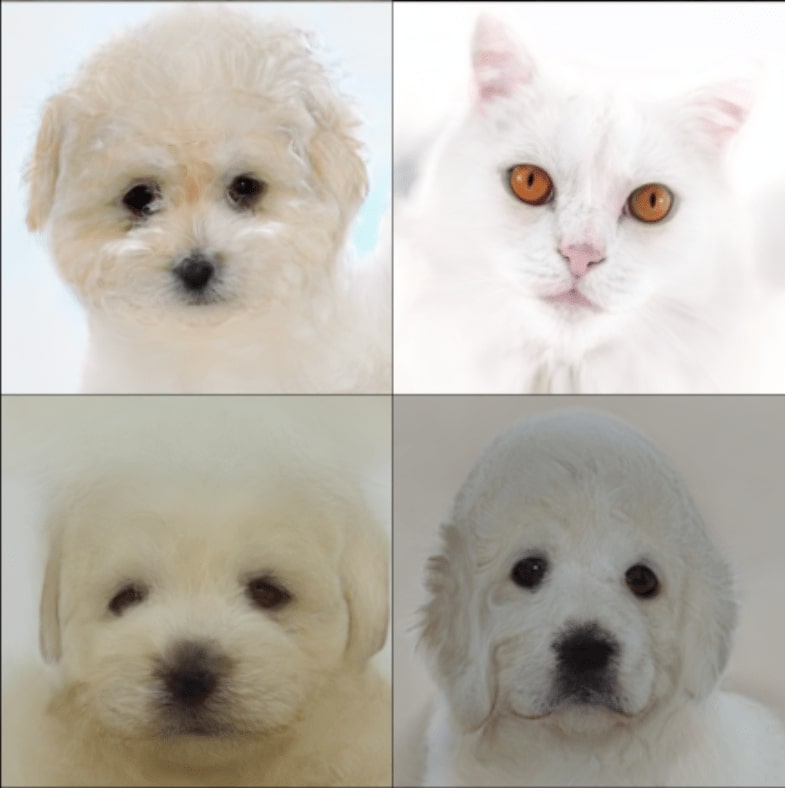}
\begin{minipage}{\columnwidth} \centering \vspace{-1em} {\tiny StyleGAN3 AFHQv2}  \end{minipage}
\end{minipage}
\begin{minipage}{0.32\linewidth}
\begin{minipage}{\columnwidth} \centering {\tiny BigGAN Egyptian cat}  \end{minipage}
\includegraphics[width=.99\columnwidth]{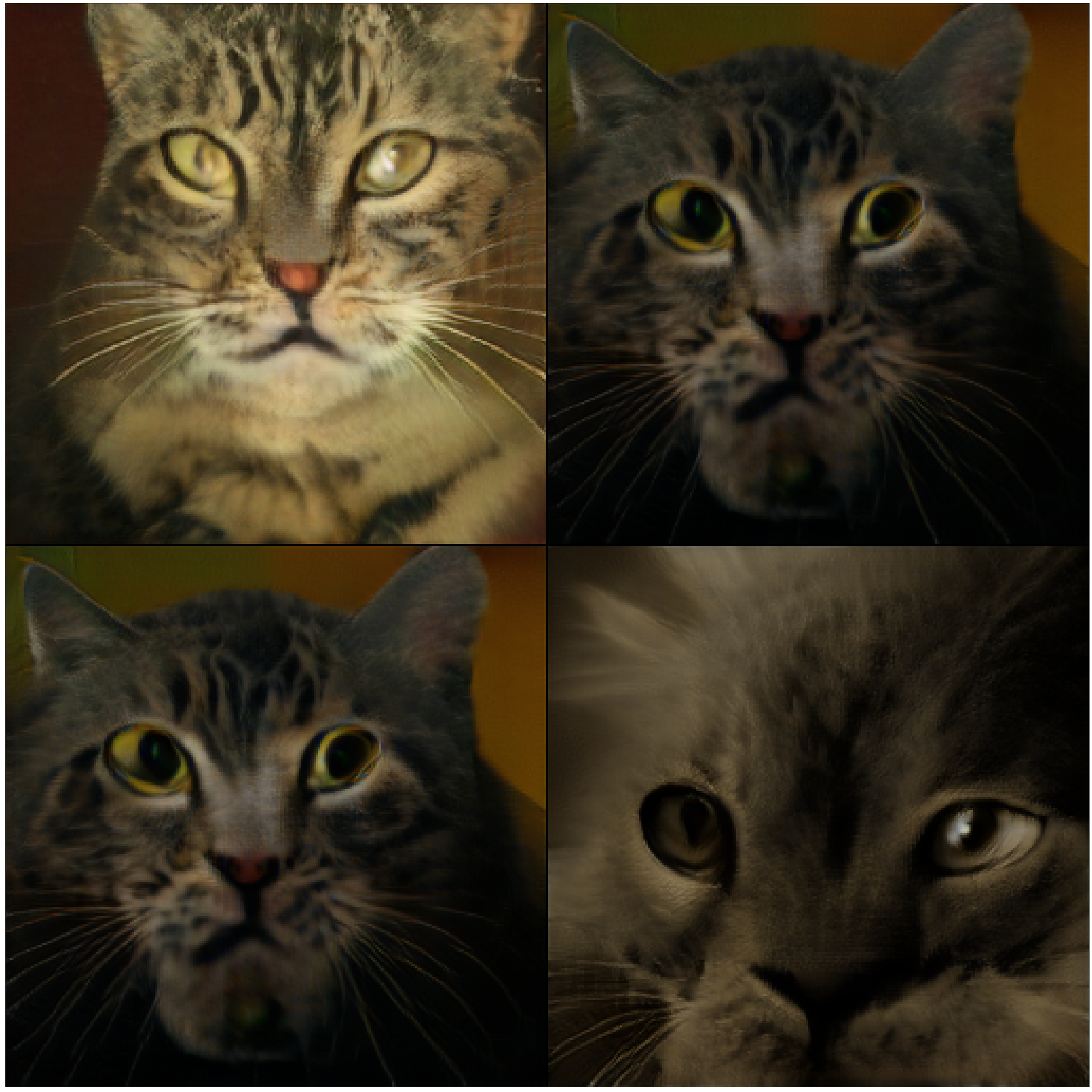}\\
\includegraphics[width=.99\columnwidth]{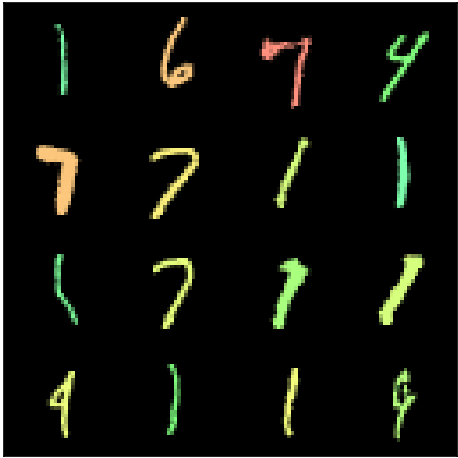}
\begin{minipage}{\columnwidth} \centering \vspace{-1em} {\tiny NVAE colored-MNIST}  \end{minipage}
\end{minipage}
\end{center}
\vspace{-0.7cm}
\caption{Modes for BigGAN-deep, StyleGAN3-T and NVAE obtained via $\rho\ll0$ Polarity Sampling. {\bf This is, to the best of our knowledge, the first visualization of the modes of DGNs in pixel space.}}
\label{fig:modes}
\end{figure}

\begin{figure}[t!]
\begin{center}
\begin{minipage}{0.49\linewidth}
\includegraphics[width=.99\columnwidth]{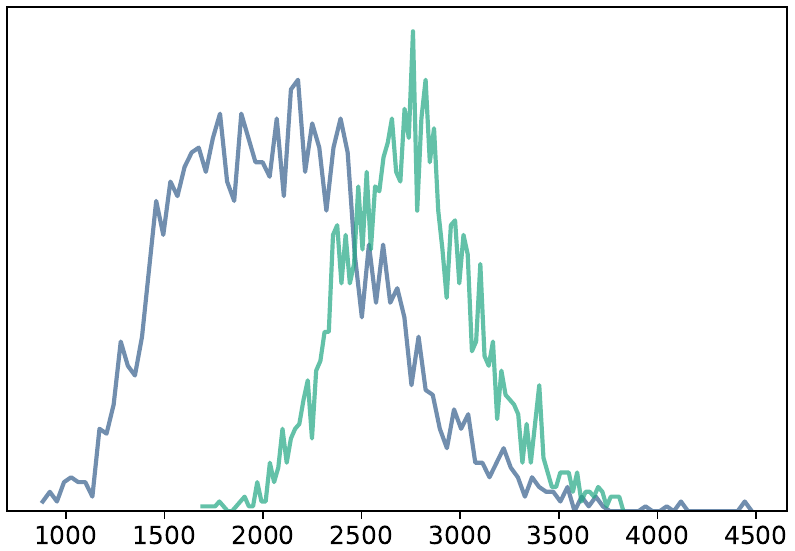}
\end{minipage}
\begin{minipage}{0.49\linewidth}
\includegraphics[width=.99\columnwidth]{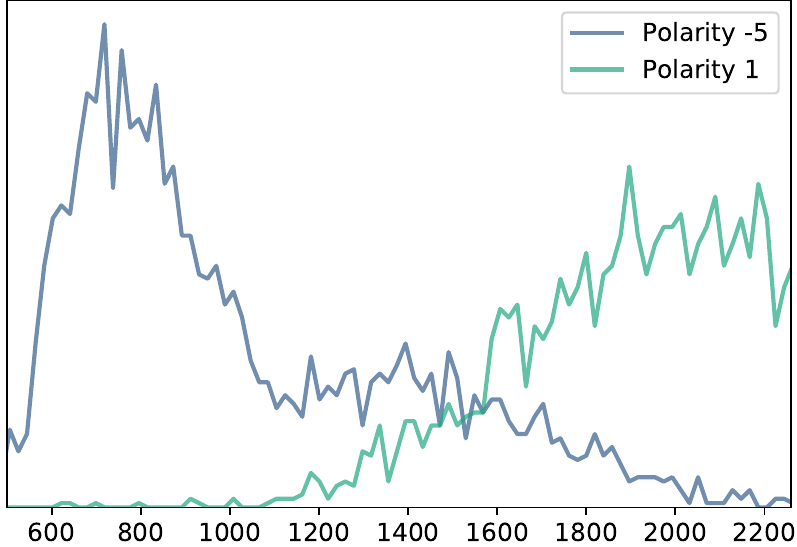}
\end{minipage}
\end{center}
\vspace{-0.5cm}
\caption{Distribution of $l_2$ distance to 3 training set nearest neighbors at $32\times32$ resolution, for 1000 generated samples from LSUN Church StyleGAN2 {\bf (left)} and colored-MNIST NVAE {\bf (right)}. Samples closer to the modes ($\rho<0$) have a significant shift in the distribution closer to the training samples for NVAE, while for StyleGAN2 the distribution shift is minimal with significant overlap. This behavior is expected as {\bf VAE models are encouraged to position their modes on the training samples, as opposed to GANs whose modes depend on the discriminator}.
}
\label{fig:nn_hist}
\end{figure}

\vspace{-0.2cm}
Mode collapse \cite{metz2016unrolled,srivastava2017veegan,bang2021mggan} has complicated GAN training for many years. It consists of the entire DGN collapsing to generate a few different samples or modes. 
For VAEs, modes can be expected to be related to the modes of the empirical dataset distribution, as reconstruction is part of the objective. But this might not be the case with GANs e.g., the modes can correspond to parts of the space where the discriminator is the least good at differentiating between true and fake samples. There has been no reported methods in literature that allows us to observe the modes of a trained GAN. Existing visualization techniques focus on finding the role of each DGN unit \cite{bau2018gan} or finding images that GANs cannot generate \cite{bau2019seeing}.
Using Polarity Sampling, we can visualize the modes of DGNs for the first time. In \cref{fig:modes}, we present samples from the modes of BigGAN-deep trained on ImageNet, StyleGAN3 trained on AFHQv2, and NVAE trained on colored-MNIST. We observe that BigGAN modes tend to reproduce the unique features of the class, removing the background and focusing more on the object that the class is assigned to. AFHQv2 modes on the other hand, focus on younger animal faces and smoother textures. NVAE mode sampling predominately produce the digit `1' which corresponds to the dataset mode (digit with the least intra-class variation).
We also provide in \cref{fig:nn_hist} the distribution of the $l_2$ distances between generated samples and their $3$ nearest training samples for modal ($\rho=$ $-5$) and anti-modal ($\rho=1$) polarity. We see that even after reducing the polarity, StyleGAN2 nearest neighbor distributions have overlap whereas for NVAE the modes move significantly closer to the training samples. In Appendix.~\cref{fig:nn_hist_mnist} we observe a similar effect for WGAN and NVAE trained on MNIST.

\begin{figure}[t!]
\begin{center}
\begin{minipage}{0.49\linewidth}
\includegraphics[width=.99\columnwidth]{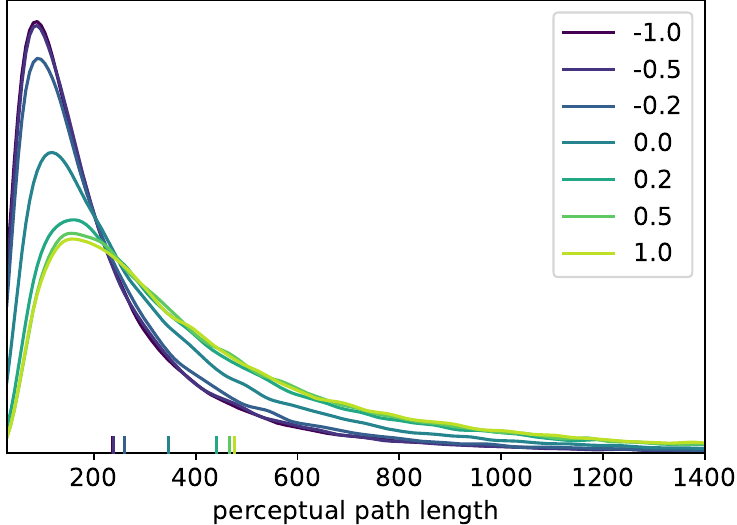}
\end{minipage}
\begin{minipage}{0.49\linewidth}
\includegraphics[width=.99\columnwidth]{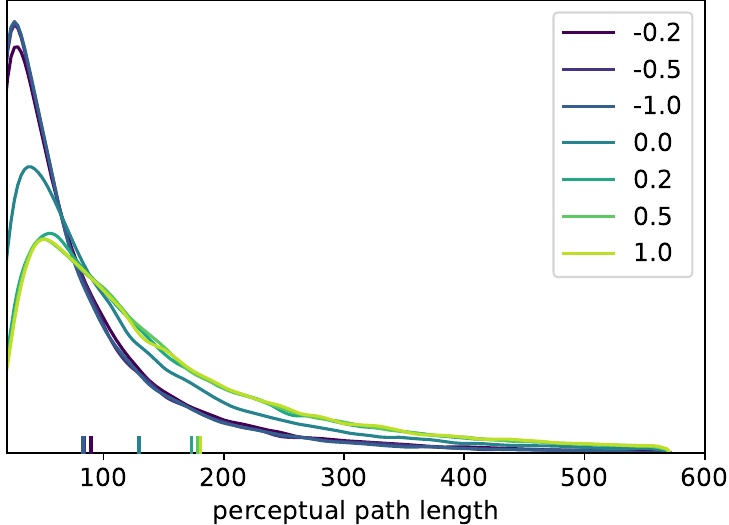}
\end{minipage}
\end{center}
\vspace{-0.5cm}
\caption{Distribution of PPL for StyleGAN2-F trained on FFHQ with varying Polarity Sampling (in VGG space) setting ($\rho$ given in the legend) for endpoints in the input latent space ({\bf left}) and endpoints in style-space ({\bf right}). The means of the distributions (PPL score) are provided as markers on the horizontal axis.
}
\label{fig:PPL}
\end{figure}


\subsection{Perceptual Path Length Around Modes}
\label{sec:PPL}
Perceptual Path Length (PPL) is the distance between the Vgg space image of two latent space points. It has previously been proposed as a measure of perceptual distance \cite{karras2020analyzing}. In \cref{fig:PPL}, we report the PPL of a StyleGAN2-F trained on FFHQ, for an interpolation step of length $10^{-4}$ between endpoints from the latent/style space. We sample points using Polarity Sampling varying $\rho \in [1,-1]$, essentially measuring the PPL for regions of the data manifold with increasing density as we increase $\rho$. We see that for negative values of polarity, we have significantly lower PPL compared to positive polarity or even baseline sampling ($\rho=0$). This result shows that for StyleGAN2, there are smoother perceptual transitions closer to modes. 
While truncation also reduces the PPL, it essentially does so by sampling points closer to the style space mean \cite{karras2019style}, see \cref{appendix:ppl_fixed} for comparisons. Polarity Sampling in the Vgg space, can be used to directly sample from Vgg modes, making it the first method that can be used to explicitly sample regions that are perceptually smoother. It can therefore be used to develop sophisticated interpolation methods where, the interpolation is done along a high-likelihood path on a feature space manifold.
\section{Conclusions}
We have proposed a new  parameterization of the DGN prior $p_{\vz}$ in terms of a single parameter -- the polarity $\rho$-- to force the DGN samples to be concentrated on the distribution modes or anti-modes (\cref{sec:method}). As a byproduct, for a range of DGNs, we improve the state-of-the-art FID performance.
On the theoretical side, Polarity Sampling's guarantee that it samples from the modes of a DGN enabled us to explore some timely open questions,
including the relation between distribution modes and training samples (\cref{sec:mode}), and the effect of going from mode to anti-mode generation on the perceptual path length (\cref{sec:PPL}). We show that Polarity sampling can also be performed on feature space distributions of classifiers appended with a generator, which can be possibly used for fair attribute generation, out-of-distribution synthetic data generation and much more.

\vspace{-0.2cm}
\section*{Acknowledgements}
\vspace{-0.2cm}
{\small Humayun and Baraniuk were supported by NSF grants CCF-1911094, IIS-1838177, and IIS-1730574; ONR grants N00014-18-12571, N00014-20-1-2534, and MURI N00014-20-1-2787; AFOSR grant FA9550-22-1-0060; and a Vannevar Bush Faculty Fellowship, ONR grant N00014-18-1-2047.}
\newpage

{
    \small
    \bibliographystyle{ieee_fullname}
    \bibliography{main}
}

\clearpage
\clearpage
\appendix

\setcounter{page}{1}

\twocolumn[
\centering
\Large
\textbf{Polarity Sampling: Quality and Diversity Control of Pre-Trained Generative Networks via Singular Values} \\
\vspace{0.5em}Supplementary Materials \\
\vspace{.5em}
{\small  Codes available at \href{https://bit.ly/magnet-polarity}{github.com/AhmedImtiazPrio/magnet-polarity}\\Google Colab demo
\href{https://bit.ly/polarity-samp}{bit.ly/polarity-samp}\\

}
\vspace{.5em}
] 
\appendix

We provide the following supplementary materials (SMs) as support of our theoretical and empirical claims. This SM is organized as follows. 

\cref{sec:implementation} provides all the implementation details. We first provide the pseudocode for the online sampling algorithm in \cref{sec:online_algorithm} that allows for polarity sampling to be performed rapidly for real-time applications. Along it, we provide further details on the effect of the two hyper-parameters of polarity sampling that are $N$ and $k$, namely the number of samples used to estimate the per-region singular values of $\mA_{\omega}$ for as many $\omega$ i.e. $N$ of them, and the number $k$ of top-singular values to utilize (\cref{sec:effect_N_k}). We then describe the computation times we observed on our hardware/software (\cref{sec:times}).

\cref{sec:proofs} provides the proofs for \cref{thm:general_density} and \cref{cor:pol_density}. \cref{appendix:extra} supports our claims with additional experiments on various dataset and models. First, \cref{sec:shift} studies how polarity sampling can help under distribution shift between the training distribution and a target distribution, this is done through colored-MNIST and NVAE. Then \cref{sec:progan} proposes to study the effect of polarity sampling on ProGAN which is crucial as ProGAN does not allow for truncation based control of its samples.

We conclude with \cref{sec:datasets} and \cref{sec:quality} that provide descriptions of the datasets and additional qualitative samples from the empirical experiments performed in the main part of the paper, respectively.

\section{Implementation Details and Online Sampling Solution}
\label{sec:implementation}

\subsection{Online Algorithm}
\label{sec:online_algorithm}

One important aspect of polarity sampling, as summarized in \cref{alg:sampling} is the need to first sample the DGN latent space to obtain the top singular values of as many per-region slope matrices $\mA_{\omega}$ as possible. This might seem as a bottleneck if one wants to repeatedly apply polarity sampling on a same DGN. However, this is to provide an estimate of the DGN per-region change of variables, and as the DGN is not retrained nor fine-tuned, it only needs to be done once. Furthermore, this allows for an online sampling algorithm that we provide in \cref{alg:online}. In short, one first perform this task of estimating as many per-region top singular values as possible, once this is completed, only sampling of latent vectors $\vz$ and rejection sampling based on the corresponding $\mA_{\omega}$ matrix is done online, $\mA_{\omega}$ of the sampled $\vz$ being obtained easily via $\mA_{\omega}=\mJ G (\vz)$.

\begin{algorithm}[]
\caption{Online Rejection Sampling Algorithm}
\begin{algorithmic}
\Require Latent space domain, $\mathcal{D}$; Generator $\mG$; $N$ change of volume scalars $\{\sigma_1, \sigma_2,...,\sigma_N \}$; Number of singular values $K$;
\While{True}{}
\State $z \sim U(\mathcal{D})$
\State $\alpha \sim U[0,1]$
\State $\mA = \mJ_{\mG}(z)$
\State $\sigma_z = \prod^K_{k=1}  K$-$SingularValues(\mA,K)$
\vspace{.5em}
\State \algorithmicif $\frac{\sigma_z^\rho}{\sigma_z^\rho + \sum_{i=1}^N \sigma_i^\rho}  \geq \alpha$ \algorithmicthen\
\vspace{.5em}

$x \gets \mG(z)$

\Return $x$
\EndWhile

\end{algorithmic}
\label{alg:online}
\end{algorithm}


\subsection{Effect of $N$ and $k$}
\label{sec:effect_N_k}

\begin{figure}[t!]
\begin{center}
\begin{minipage}{0.02\linewidth}
\centering
\rotatebox{90}{\footnotesize \hspace{0.7cm}Precision}
\end{minipage}
\begin{minipage}{0.48\linewidth}
\centering
\includegraphics[width=\linewidth]{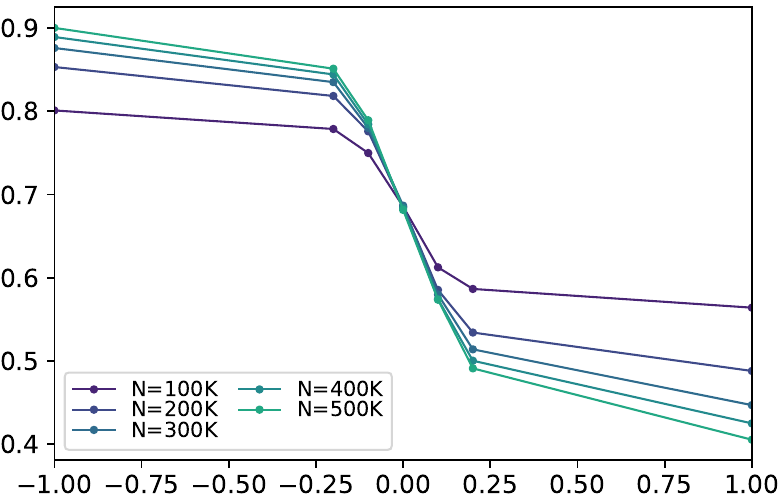}
\end{minipage}
\begin{minipage}{0.48\linewidth}
\centering
\includegraphics[width=\linewidth]{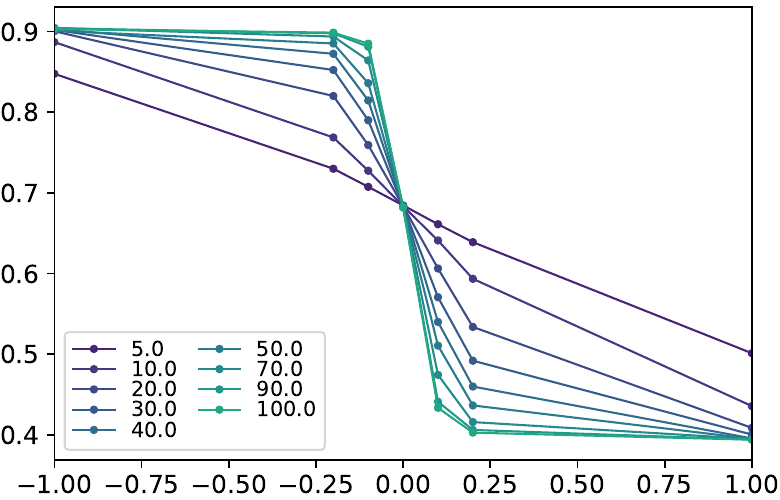}
\end{minipage}
\begin{minipage}{0.02\linewidth}
\centering
\rotatebox{90}{\footnotesize \hspace{0.7cm}Recall}
\end{minipage}
\begin{minipage}{0.48\linewidth}
\centering
\includegraphics[width=\linewidth]{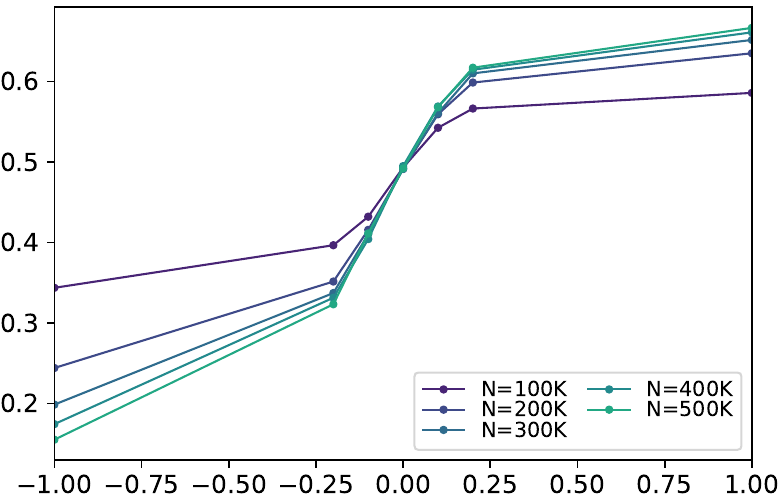}\\[-0.5em]
{\footnotesize polarity $(\rho)$}
\end{minipage}
\begin{minipage}{0.48\linewidth}
\centering
\includegraphics[width=\linewidth]{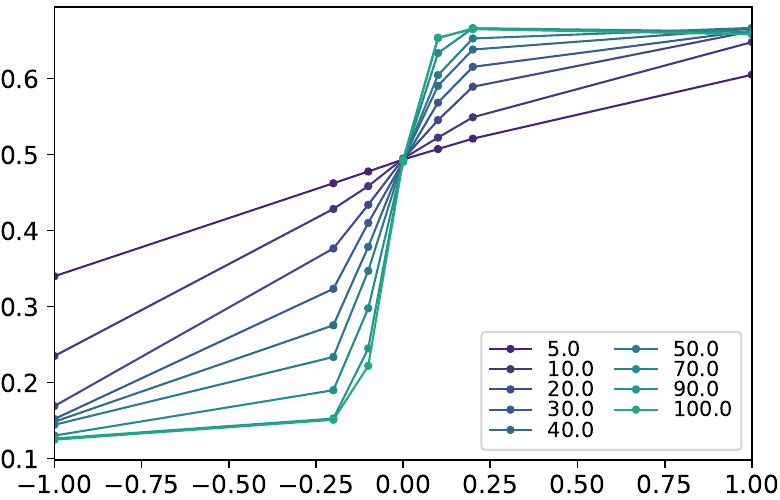}\\[-0.5em]
{\footnotesize polarity $(\rho)$}
\end{minipage}
\end{center}
\vspace{-0.5cm}
\caption{ Effect of Polarity Sampling on Precision (top) and Recall (bottom) of a StyleGAN2-F model pretrained on FFHQ for varying number of top-$k$ singular values (right) and varying number of latent space samples $N$ (left) used to obtain per-region slope matrix $\mA_{\omega}$ singular values (recall \cref{sec:pseudocode,alg:sampling}). The trend in metrics  stabilizes when using around $N\approx$300,000 latent space samples. Increasing the number of top-$k$ singular values to use, amplifies the effect of polarity, saturating at around $k=50$.
}
\label{fig:topk_prec_recall}
\end{figure}

One important aspect of our algorithm comes from the two hyper-parameters $N$ and $k$. They represent respectively the number of latent space samples to use to estimate as much $\mA_{\omega}$ as possible (recall \cref{alg:sampling}), and the number of top singular values to compute. Both represent a trade-off between exact polarity sampling, and computation complexity. We argue that in practice, $N\approx 150K$ and $k\approx 100$ is enough to obtain a good estimate of the polarity sampling distribution (\cref{eq:density_rho}). To demonstrate that, we first provide an ablation study of the number of $N$ and $k$ used for polarity sampling in \cref{tab:ablation_N} and \cref{tab:ablation_topk}. We also present a visual inspection of the impact of $N$ and $k$ on the precision and recall in \cref{fig:topk_prec_recall}.

\begin{table}[]
\centering
\resizebox{\linewidth}{!}{%
\begin{tabular}{@{}crrrccc@{}}
\toprule
 & \multicolumn{3}{c}{FFHQ 1024$\times$1024} & \multicolumn{3}{c}{LSUN Cat 256$\times$256} \\ \midrule
$N$ & \multicolumn{1}{c}{\begin{tabular}[c]{@{}c@{}}FID\\ (lowest)\end{tabular}} & \multicolumn{1}{c}{\begin{tabular}[c]{@{}c@{}}Precision\\ (max)\end{tabular}} & \multicolumn{1}{c}{\begin{tabular}[c]{@{}c@{}}Recall\\ (max)\end{tabular}} & \begin{tabular}[c]{@{}c@{}}FID\\ (lowest)\end{tabular} & \begin{tabular}[c]{@{}c@{}}Precision\\ (max)\end{tabular} & \begin{tabular}[c]{@{}c@{}}Recall\\ (max)\end{tabular} \\ \midrule
100K & 2.63 & 0.80 & 0.59 & 6.38 & 0.69 & 0.31 \\
200K & 2.62 & 0.82 & 0.63 & 6.38 & 0.71 & 0.32 \\
250K & 2.59 & 0.84 & 0.64 & 6.39 & 0.74 & 0.31 \\
300K & 2.61 & 0.87 & 0.65 & 6.37 & 0.75 & 0.33 \\
500K & 2.58 & 0.90 & 0.67 & 6.34 & 0.77 & 0.33
\end{tabular}
}
\caption{Ablation of $N$ and its effect on best FID, Precision and Recall values that can be obtained by a StyleGAN2 ($\psi=1$) on the FFHQ and LSUN Cat dataset. We vary the polarity in the VGG space for FFHQ dataset, and style space for LSUN Cats, for number of singular values $k=30$.}
\label{tab:ablation_N}
\end{table}

    

\begin{table}[]
\centering
\begin{tabular}{@{}crrr@{}}
\toprule
 & \multicolumn{3}{c}{FFHQ 1024$\times$1024} \\ \midrule
$K$ & \multicolumn{1}{c}{\begin{tabular}[c]{@{}c@{}}FID\\ (lowest)\end{tabular}} & \multicolumn{1}{c}{\begin{tabular}[c]{@{}c@{}}Precision\\ (max)\end{tabular}} & \multicolumn{1}{c}{\begin{tabular}[c]{@{}c@{}}Recall\\ (max)\end{tabular}} \\ \midrule
10 & 2.71 & 0.89 & 0.65 \\
20 & 2.67 & 0.90 & 0.66 \\
40 & 2.57 & 0.90 & 0.67 \\
60 & 2.62 & 0.90 & 0.66 \\
80 & 2.67 & 0.90 & 0.66 \\
100 & 2.70 & 0.90 & 0.67 \\ \bottomrule
\end{tabular}
    \caption{Ablation of $K$ and its effect on the best FID, Precision and Recall values that can be obtained by a StyleGAN2 ($\psi=1$) on the FFHQ dataset. We vary the polarity in the inception space for FFHQ dataset}
    \label{tab:ablation_topk}
\end{table}



\subsection{Computation Times and Employed Software/Hardware}
\label{sec:times}

All the experiments were run on a Quadro RTX 8000 GPU, which has 48 GB of high-speed GDDR6 memory and 576 Tensor cores. For the software details we refer the reader to the provided codebase. In short, we employed TF2 (2.4 at the time of writing), all the usual Python scientific libraries such as NumPy and PyTorch. We employed the official repositories of the various models we employed with official pre-trained weights.
As a note, most of the architectures can not be run on GPUs with less or equal to 12 GB of memory.

We report here the Jacobian computation times for Tensorflow 2.5 with CUDA 11 and Cudnn 8 on an NVIDIA Titan RTX GPU. For StyleGAN2 pixel space, 5.03s/it; StyleGAN2 style-space, 1.12s/it; BigGAN 5.95s/it; ProgGAN 3.02s/it. For NVAE on Torch 1.6 it takes 20.3s/it. Singular value calculation for StyleGAN2 pixel space takes .005s/it, StyleGAN2 style space .008s/it, BigGAN .001s/it, ProgGAN .004s/it and NVAE .02s/it on NumPy. According to this, for StyleGAN2-e, N=250,000 requires ~14 days to obtain. This only needs to be done once, and it is also possible to perform online sampling once it is calculated. The time required for this is relatively small compared to the training time required for only one set of hyperparameters, which is ~35 days and 11 hours\footnote{https://github.com/NVlabs/stylegan2}. We have added pseudocode for MaGNET sampling and online sampling in Appendix G.

{\bf Computational Complexity.}~We are computing the top-$k$ singular values of the $D \times K$ Jacobian matrix. This can be performed in $\mathcal{O}(DKk+Dk^2)$. In fact, one has to project $k$ $K$-dimensional vectors onto the $D\times K$ Jacobian's matrix: $\mathcal{O}(DKk)$ and then perform QR-decomposition of the $D\times k$ matrix: $\mathcal{O}(Dk^2)$. Then, $k$ $D$-dimensional vectors are projected onto the transpose of the Jacobian matrix: $\mathcal{O}(DKk)$ followed by their QR-decomposition: $\mathcal{O}(Kk^2)$, dominated by $\mathcal{O}(Dk^2)$ (full SVD runs in $\mathcal{O}(DK^2)$). 

\subsection{Reducing Memory Requirements}
\label{sec:memory}

The core of polarity sampling relies on computing the top-singular values of the possibly large matrix $\mA_{\omega}$ for a variety of regions $\omega \in \Omega$, discovered through latent space sampling (recall \cref{alg:sampling}). One challenge for state-of-the-art DGNs lies in the size of the matrices $\mA_{\omega}$. Multiple solutions exist, such as computing the top singular values through block power iterations. Doing so, the matrices $\mA_{\omega}$ do not need to be computed entirely, only the matrix-matrix product $\mA_{\omega}\mW$ and $\mA^T_{\omega}\mV$ needs to be performed repeatedly (interleaved with QR decompositions). After many iterations, $\mW$ estimate the top right-singular vectors of $\mA_{\omega}$, and $\mV$ the corresponding top left-singular vectors from which the singular values can be obtained. However, we found this solution to remain computationally extensive, and found that in practice, a simpler approximation that we now describe provided sufficiently accurate estimates.

Instead of the above iterative estimation, one can instead compute the top-singular values of $\mW\mA_{\omega}$ with $\mW$ a semi-orthogonal matrix of shape $D'\times D$ with $D'<D$ (recall that $\mA_{\omega}$ is of shape $D \times K$). Doing so, we are now focusing on the singular values of $\mA_{\omega}$ whose left-singular vectors are not orthogonal with the right singular vectors of $\mW$. While this possibly incurs an approximation error, we found that the above was sufficient to provide polarity sampling and adequate precision-recall control.

\subsection{Applying Polarity Sampling in Style, VGG and Inception Space}
\label{sec:other_space}
\begin{figure}[t!]
\begin{center}
\begin{minipage}{0.02\linewidth}
\centering
\rotatebox{90}{\footnotesize \hspace{0.3cm}Precision/Recall}
\end{minipage}
\begin{minipage}{0.48\linewidth}
\centering
\includegraphics[width=\linewidth]{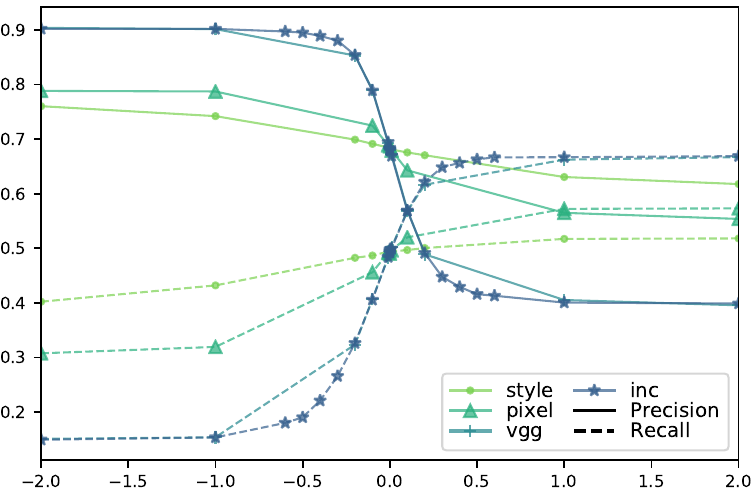}\\[-0.5em]
{\footnotesize polarity $(\rho)$}
\end{minipage}
\begin{minipage}{0.48\linewidth}
\centering
\includegraphics[width=\linewidth]{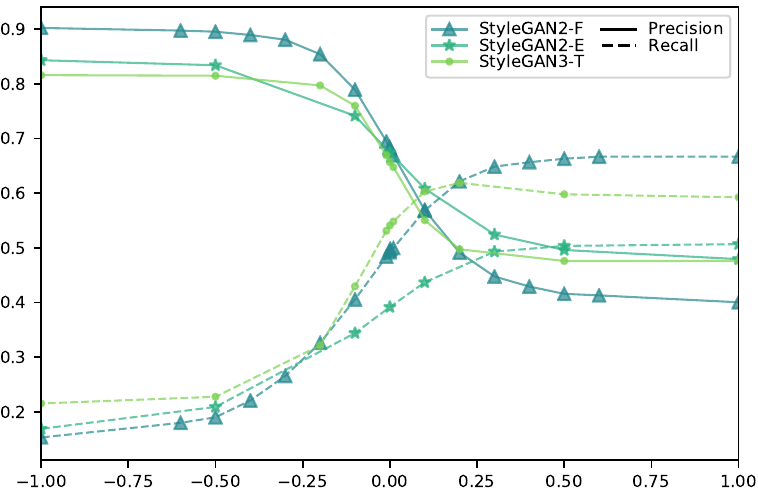}\\[-0.5em]
{\footnotesize polarity $(\rho)$}
\end{minipage}
\end{center}
\vspace{-0.5cm}
\includegraphics[width=0.49\linewidth]{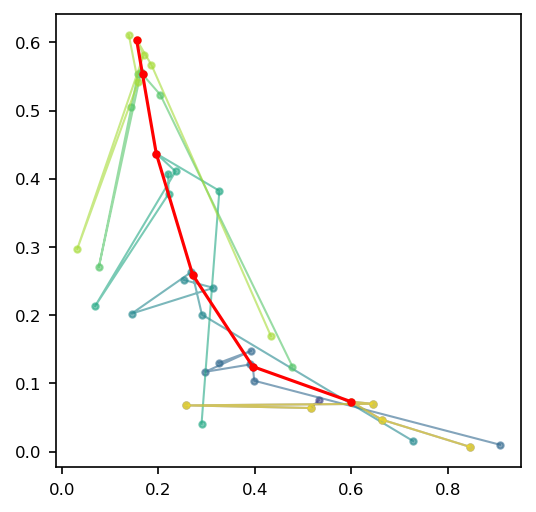}
\includegraphics[width=0.49\linewidth]{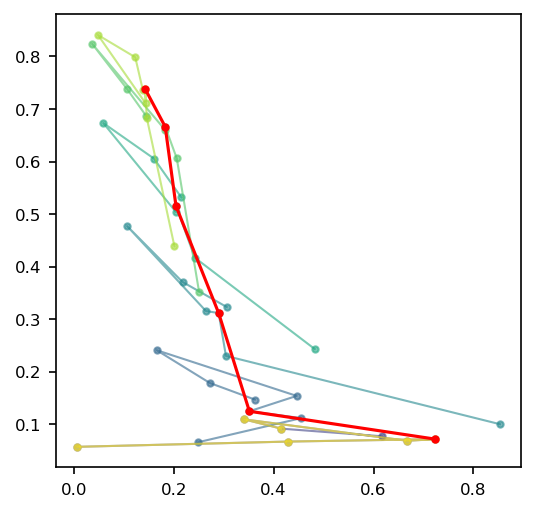}
\caption{ Top: Precision  Recall tradeoff for polarity sweep on VGG, Inception and Pixel space distributions.
Bottom: BigGAN-deep Imagenet pareto curves obtained for a few classes, in red is the baseline while each scatter point can be reached by varying truncation and $\rho$. Calculated with $1300$ real and generated samples.
}
\label{fig:more_pareto}
\end{figure}

We call the ambient space of the images the pixel-space, because each dimension in this space corresponds to individual pixels of the images. Apart from controlling the density of the pixel-space manifold, polarity can also be used to control the density of the style-space manifold for style based architectures such as StyleGAN\{1,2,3\} \cite{karras2019style,karras2020analyzing,karras2021alias}. We also extend the idea of intermediate manifolds to feature space manifolds such as VGG or InceptionV3 space, which can be assumed continuous mappings of the pixel space to the corresponding models' bottleneck embedding space. In \cref{fig:more_pareto}-left we present comparisons between Style, Pixel, VGG and Inception space precision-recall curves for StyleGAN2-F FFHQ with $\psi=1$, top-$k=30$ and $\rho=[-2,2]$. We see that the VGG and InceptionV3 curves trace almost identically. This is expected behavior since both these feature spaces correspond to perceptual features, therefore the transform they induce on the pixel space distribution is almost identical. On the other hand, the pixel space distribution saturates at high polarity at almost equal values. The point of equal precision and recall for both the Inception and VGG spaces, occurs at a polarity of 0.1. It's clear from the figures that feature space polarity changes have a larger effect on precision and recall compared to pixel-space and style-space has the least effect on precision and recall. This could be due to the number of density transforms the style-space distribution undergoes until the VGG space, where precision and recall is calculated. In \cref{fig:more_pareto}-right we present the polarity characteristics for StyleGAN2-E, StyleGAN2-F and StyleGAN3. For each model, we choose the best space w.r.t the pareto frontier, VGG and Inception space for StyleGAN2-E and StyleGAN2-F, and pixel-space for StyleGAN3. Notice that StyleGAN3 exceeds the recall of the other two models for negative polarity, while matching the precision for StyleGAN2-E.


\section{Proofs}
\label{sec:proofs}

The proofs of the two main claims of the paper heavily rely on the spline form of the DGN input-output mapping from \cref{eq:CPA}. For more background on the form of the latent space partition $\Omega$, the per-region affine mappings $\mA_{\omega},\vb_{\omega},\forall \omega \in \Omega$ and further discussion on how to deal with DGN including smooth activation functions, we refer the reader to \cite{balestriero2020mad}, and in particular to \cite{humayun2021magnet} for DGN specific results. 

\subsection{Proof of \cref{thm:general_density}}
\label{proof:general_density}

\begin{proof}
We will be doing the change of variables $\vz=(\mA^T_{\omega}\mA_{\omega})^{-1}\mA_{\omega}^T(\vx - \vb_{\omega})\triangleq \mA_{\omega}^\dagger(\vx - \vb_{\omega})$, also notice that  $J_{G^{-1}}(\vx)=A^\dagger$.
First, we know that
$
P_{G(\vz)}(\vx \in w)= P_{\vz}(\vz \in G^{-1}(w))= \int_{G^{-1}(w)}p_{\vz}(\vz) d\vz
$ which is well defined based on our full rank assumptions. We then proceed by
\begin{align*}
    P_{G}(\vx \in w)=& \sum_{\omega \in \Omega}\int_{\omega \cap w} p_{\vz}(G^{-1}(\vx)) \\
    &\times \sqrt{ \det(J_{G^{-1}}(\vx)^TJ_{G^{-1}}(\vx))} d\vx\\
    =& \sum_{\omega \in \Omega}\int_{\omega \cap w}p_{\vz}(G^{-1}(\vx)) \\
    &\times \sqrt{ \det((\mA_{\omega}^{+})^T\mA_{\omega}^{+})} d\vx\\
    =& \sum_{\omega \in \Omega}\int_{\omega \cap w} p_{\vz}(G^{-1}(\vx))\frac{1}{\sqrt{\det(\mA_{\omega}^T\mA_{\omega})}} d\vx,\\
\end{align*}
where the second to third equality follows by noticing that  $\sigma_i(A^\dagger)=(\sigma_i(A))^{-1}$ which can be showed easily be replacing  $\mA_{\omega}$ with its SVD and unrolling the product of matrices.
Now considering a uniform latent distribution case on a bounded domain $U$ in the DGN latent space we obtain by substitution in the above result
\begin{align}
    p_{G}(\vx) = \frac{\sum_{\omega \in \Omega}\mathrm{1}_{\vx \in \omega}\det(\mA_{\omega}^T\mA_{\omega})^{-\frac{1}{2}}}{Vol(U)},
\end{align}
leading to the desired result.
\end{proof}

\subsection{Proof of \cref{cor:pol_density}}
\label{proof:pol_density}

\begin{proof}
The proof of this result largely relies on \cref{thm:general_density}. Taking back our previous result, we know that 
\begin{equation}
    p_{G}(\vx)=\sum_{\omega \in \Omega} p_{\vz}(G^{-1}(\vx))\mathrm{1}_{\{G^{-1}(\vx) \in \omega\}}\frac{1}{\sqrt{\det(\mA_{\omega}^T\mA_{\omega})}} d\vx.
\end{equation}
However, recall that polarity sampling leverages the prior probability given by 
\begin{equation}
    p_{\rho}(\vz) =\frac{1}{\kappa} \sum_{\omega \in \Omega}\det(\mA_{\omega}^T\mA_{\omega})^{\frac{\rho}{2}} \mathds{1}_{\{\vz \in \omega\}},
\end{equation}
which, after replacing $G^{-1}(\vx)$ with its corresponding $\vz$ becomes
\begin{equation}
    p_{G}(\vx)=\sum_{\omega \in \Omega} \frac{1}{\kappa} \frac{\det(\mA_{\omega}^T\mA_{\omega})^{\frac{\rho}{2}}}{\sqrt{\det(\mA_{\omega}^T\mA_{\omega})}}\mathds{1}_{\{\vz \in \omega'\}} d\vx,
\end{equation}
and simplifies to
\begin{equation}
    p_{G}(\vx)=\sum_{\omega \in \Omega} \frac{1}{\kappa} \det(\mA_{\omega}^T\mA_{\omega})^{\frac{\rho-1}{2}}\mathds{1}_{\{\vz \in \omega'\}} d\vx,
\end{equation}
leading to the desired result. Note that when $\rho=1$ then the density is uniform onto the DGN manifold, when $\rho=0$, one recovers the original DGN density onto the manifold, and in the extreme cases, only the region with highest or lowest probability would be sampled i.e. the modes or anti-modes.
\end{proof}

\section{Extra experiments}
\label{appendix:extra}

\begin{figure*}[t!]
\begin{center}
\begin{minipage}{\linewidth}
\includegraphics[width=.49\columnwidth]{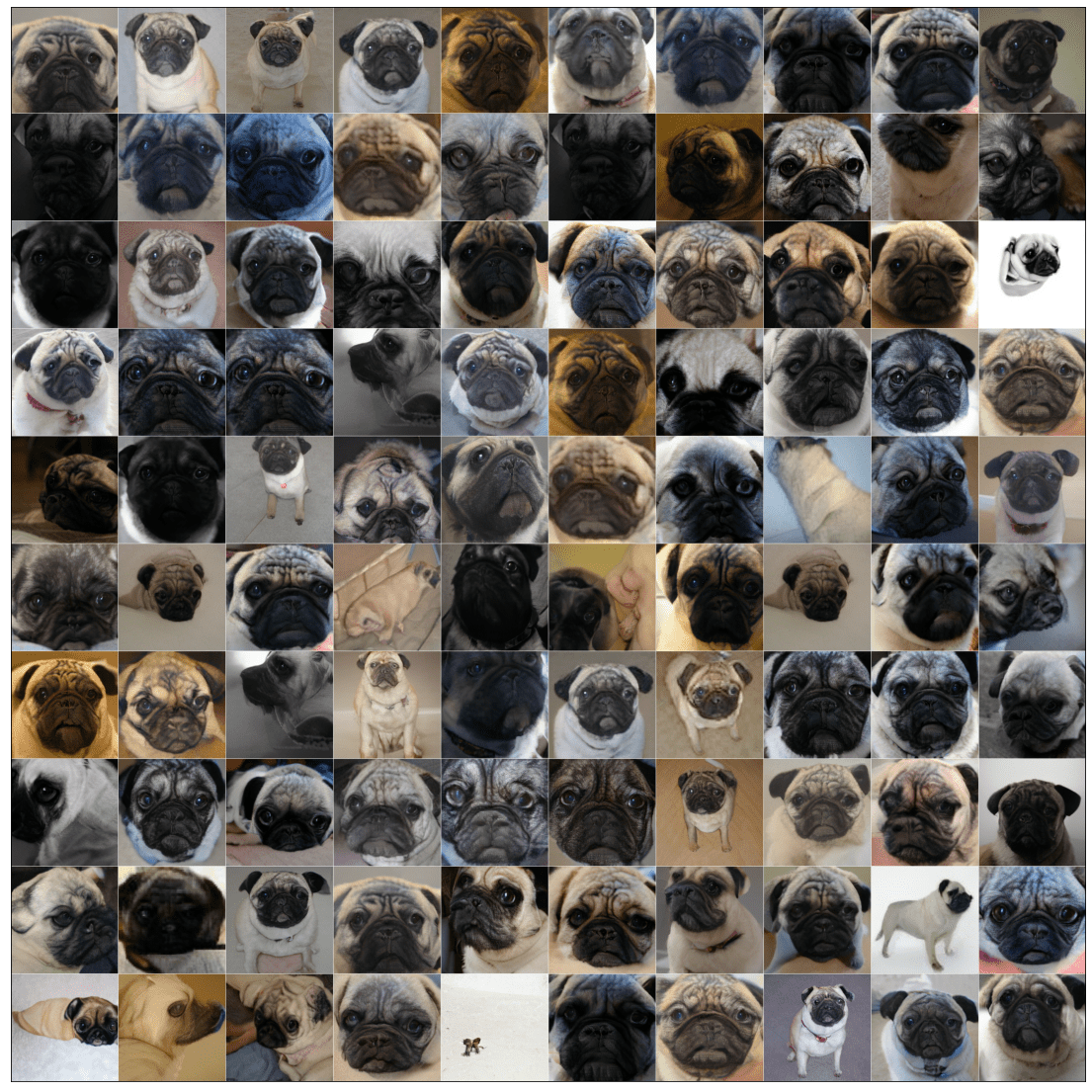}
\includegraphics[width=.49\columnwidth]{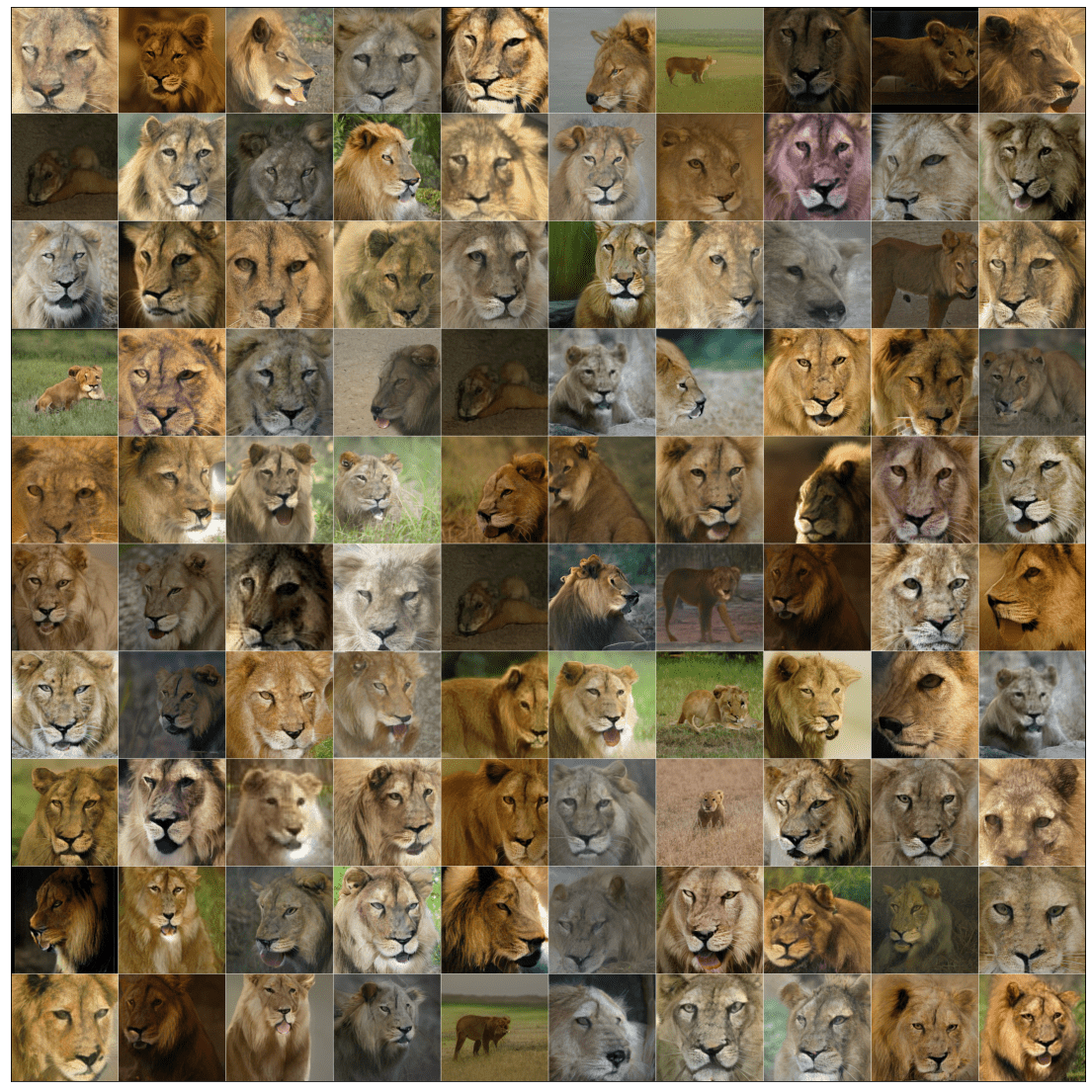}\\
\includegraphics[width=.49\columnwidth]{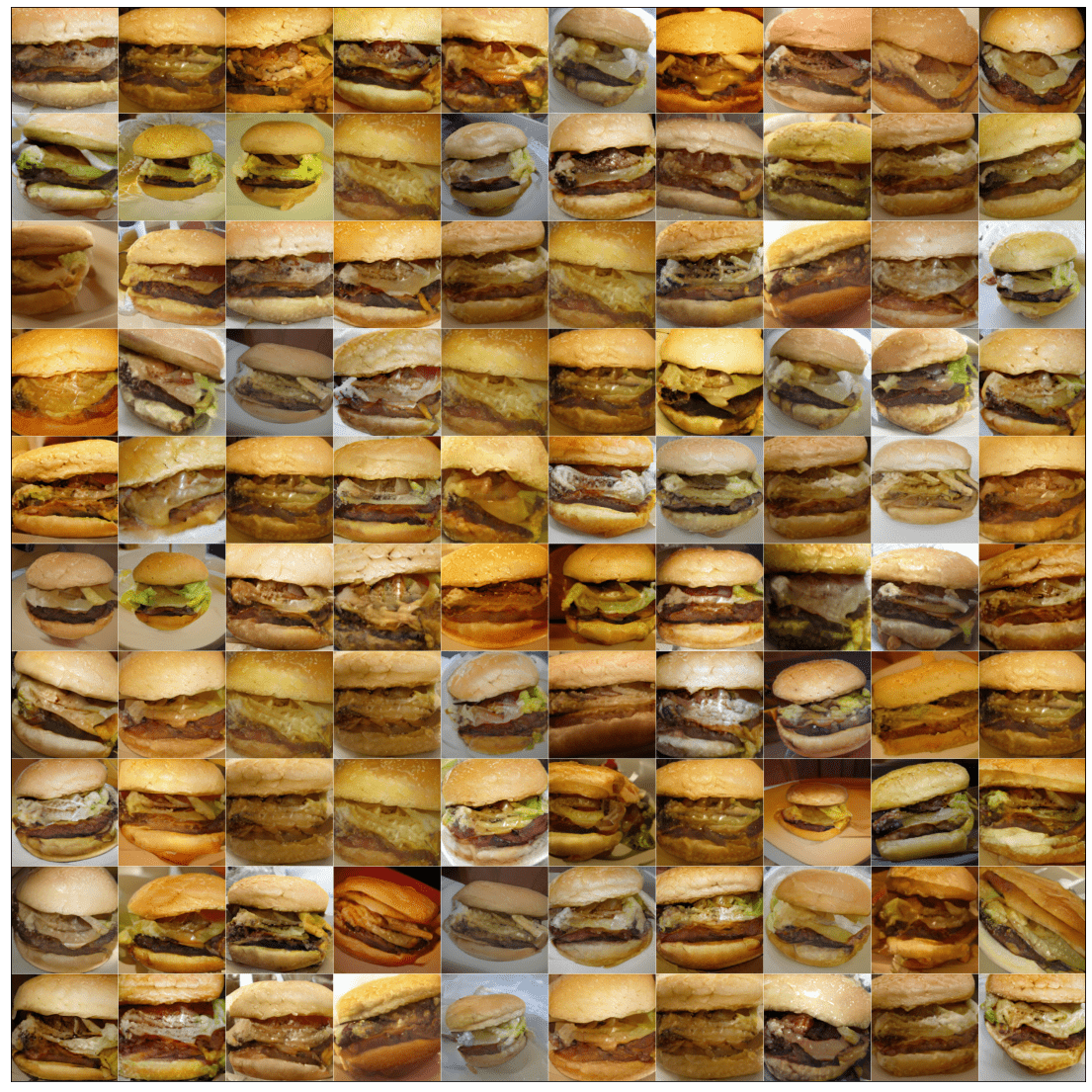}
\includegraphics[width=.49\columnwidth]{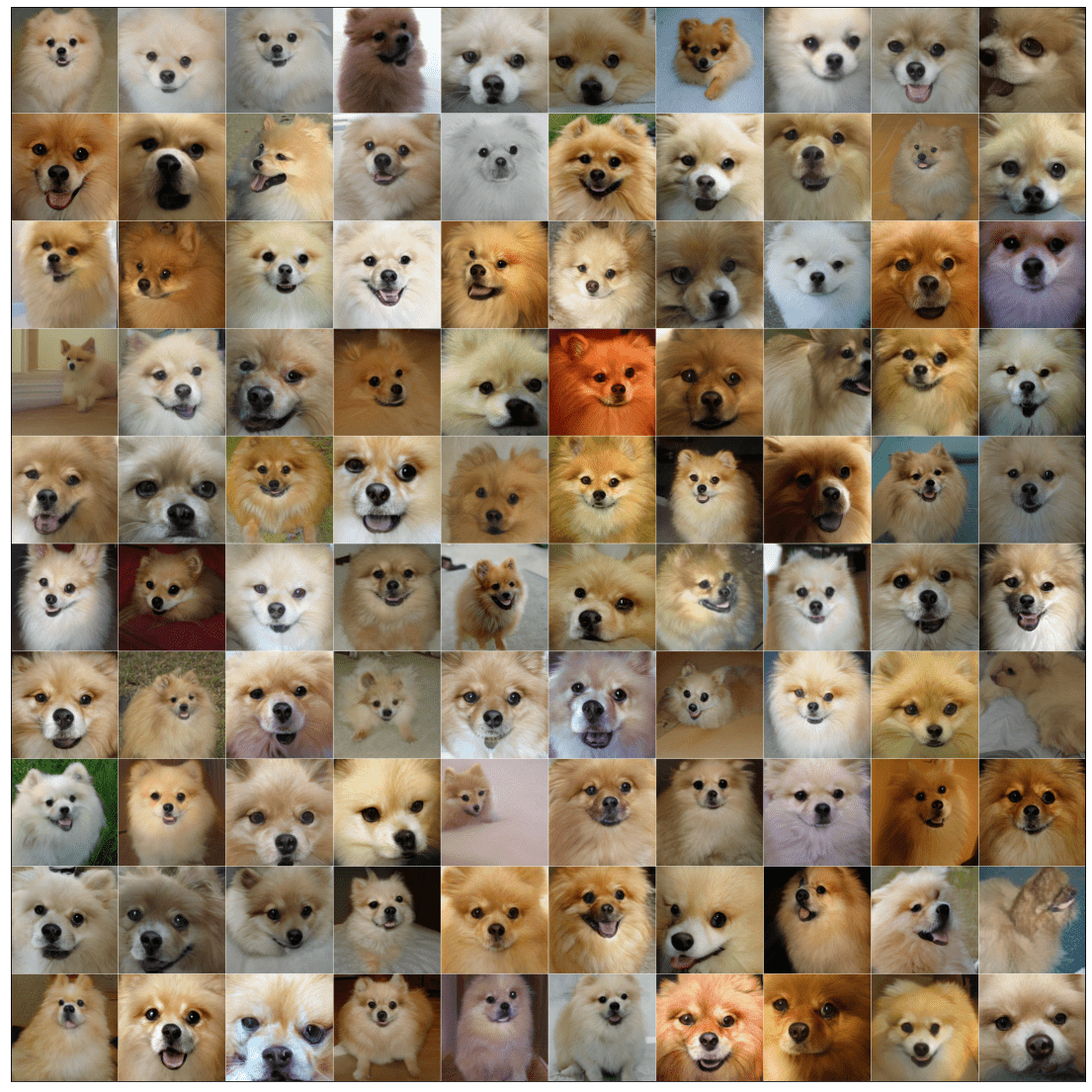}\\
\end{minipage}
\end{center}
\caption{Modes for BigGAN-deep trained on Imagenet, and conditioned on a specific class: ``pug'' (top left), ``lion'' (top right), ``cheeseburger'' (bottom left) and ``pomerian'' (bottom right). We observe that the modes correspond to nearly aligned faces with little to no background. Variation of colors and sizes can be seen across the modes. The same observation can be made for the cheeseburger, nearly no background is present, and the shape is consistent to a typical cheeseburger ''template''. See \cref{fig:modes_per_class} for additional classes.
}
\label{fig:truncation_sweep_images}
\end{figure*}

\begin{figure*}[t!]
    \centering
    \includegraphics[width=0.8\linewidth]{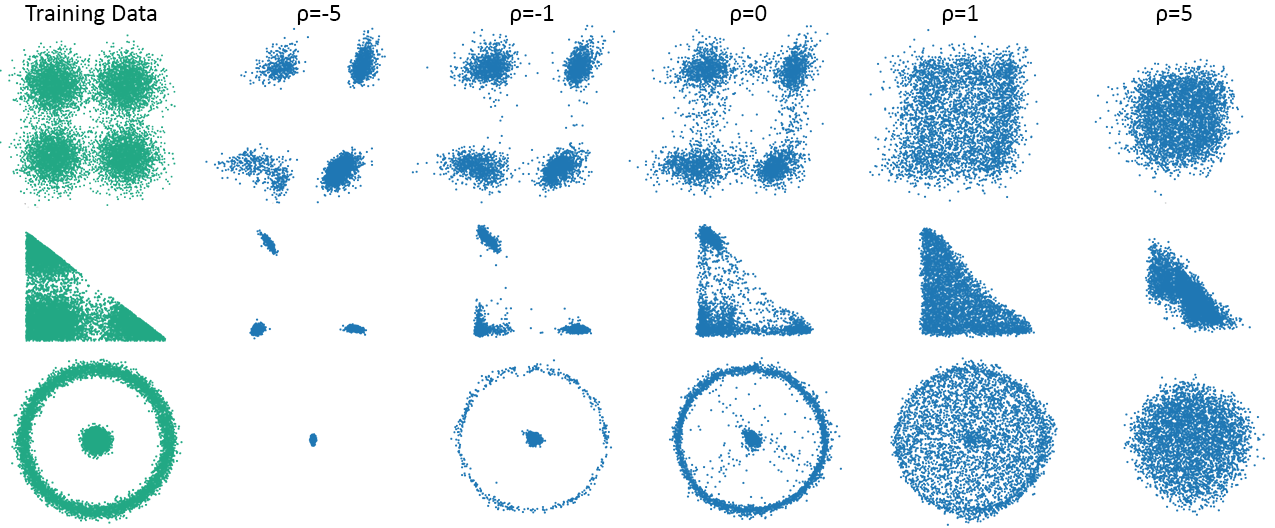}
    \caption{Polarity sweep for a WGAN trained on 2D toy datasets with $4$ gaussians (top row), $4$ gaussians with triangular domain (middle row) and two circles (bottom row).}
    \label{2d_examples}
\end{figure*}

\begin{figure}[t!]
\begin{center}
\begin{minipage}{0.48\linewidth}
\includegraphics[width=.99\columnwidth]{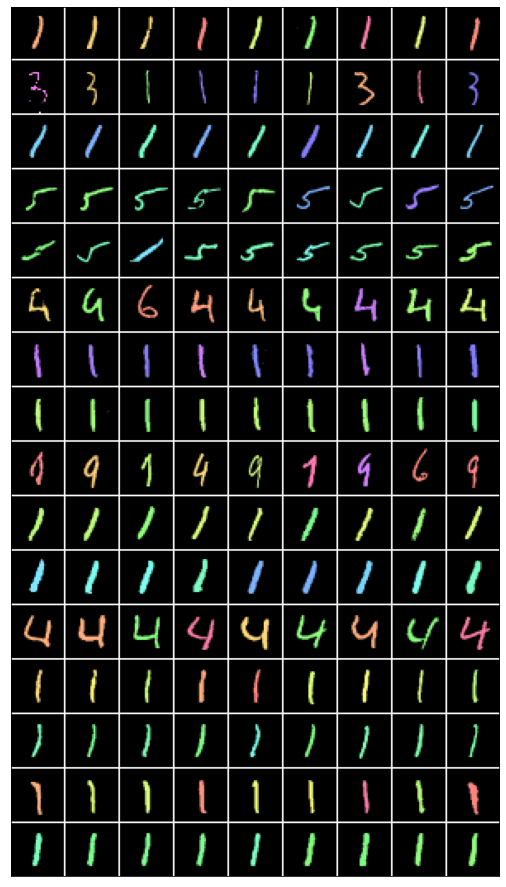}
\end{minipage}
\begin{minipage}{0.48\linewidth}
\includegraphics[width=.99\columnwidth]{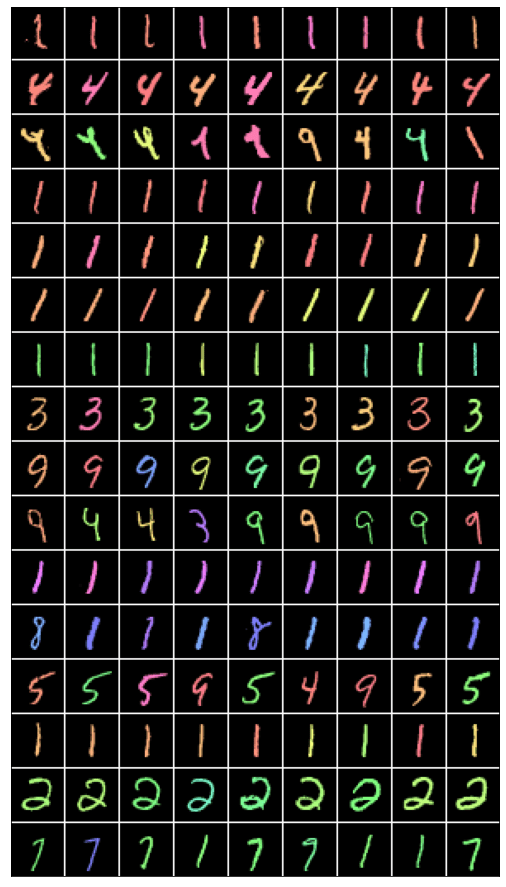}
\end{minipage}
\end{center}
\caption{Modes for a VAE trained on colored-MNIST with 8 nearest neighbors. The leftmost column for each figure contains generated samples. Notice the higher prevalence of digit 1. Due to low pixel variations, digit 1 samples have high density on the manifold.}
\label{fig:nearest}
\end{figure}

\subsection{Polarity Helps Under Distribution Shift}
\label{sec:shift}

\begin{figure}[h!]
    \centering
        \includegraphics[width=.32\linewidth]{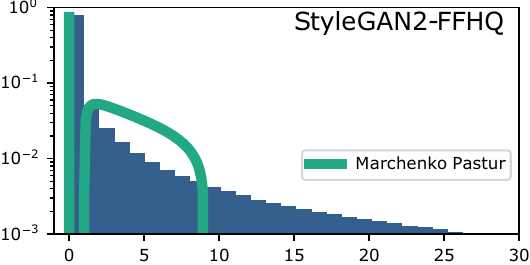}
    \includegraphics[width=.32\linewidth]{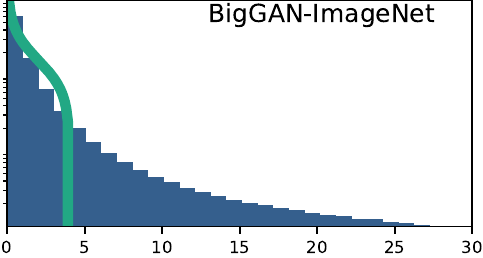}
    \includegraphics[width=.32\linewidth]{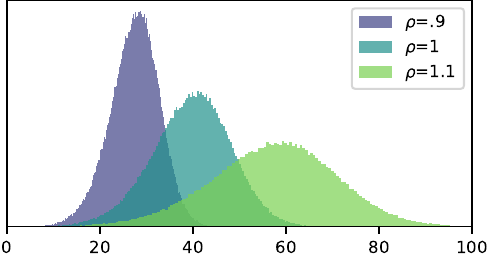}
    \caption{Singular value distribution and Marchenko Pastur distribution fit for StyleGAN2-FFHQ ({\bf left}) and BigGAN-Imagenet ({\bf middle}). Log-sigma distribution for StyleGAN2-FFHQ with varying $\rho$ ({\bf right}).}
    \label{fig:sv_hist}
\end{figure}

\begin{figure}[t!]
    \centering
    \begin{minipage}{0.49\linewidth}
    \centering
    \includegraphics[width=\linewidth]{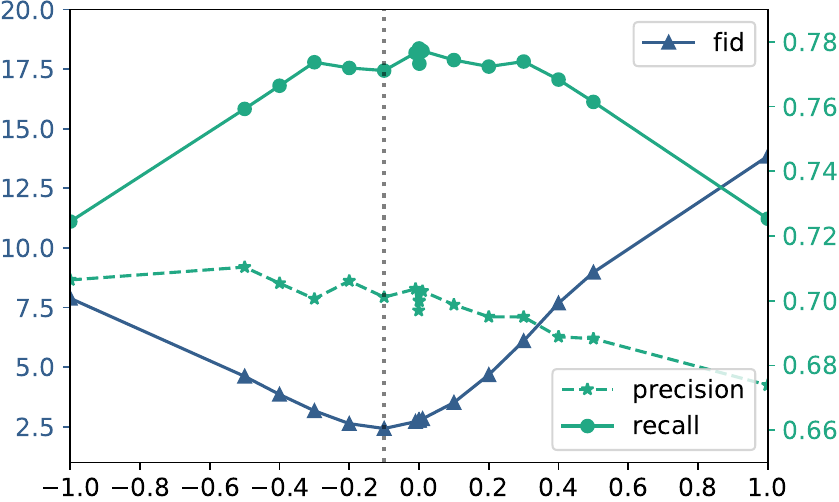}\\[-0.5em]
    {\small polarity ($\rho$)}
    \end{minipage}
    \begin{minipage}{0.49\linewidth}
    \centering
    \includegraphics[width=\linewidth]{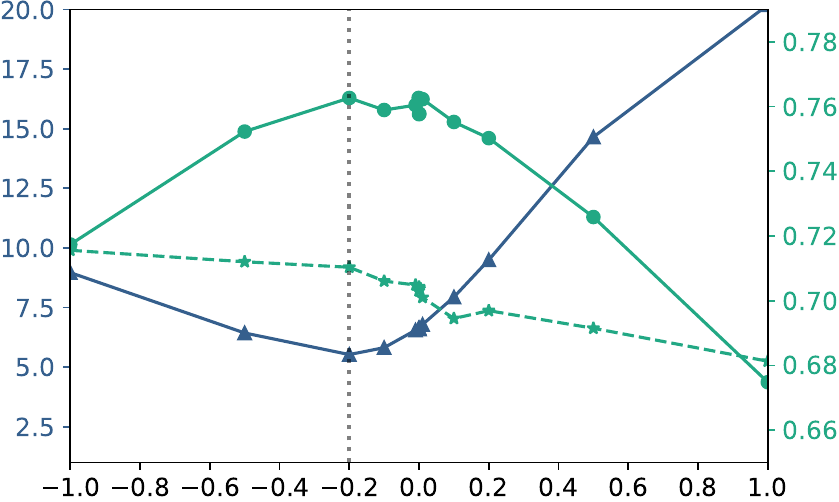}\\[-0.5em]
    {\small polarity ($\rho$)}
    \end{minipage}
    \vspace{-0.2cm}
    \caption{FID, precision and recall for an NVAE trained on colored-MNIST with hue bias. Metrics are calculated for a test dataset with hue distribution  {\bf Left:} identical to training and {\bf Right:} uniformly distributed across digit classes. {\bf Polarity allows adapting the DGN output distribution to balance possible distribution shifts}.}
    \label{fig:nvae}
\end{figure}

\begin{figure}[t!]
    \centering
    \begin{minipage}{0.49\linewidth}
    \centering
    \includegraphics[width=\linewidth]{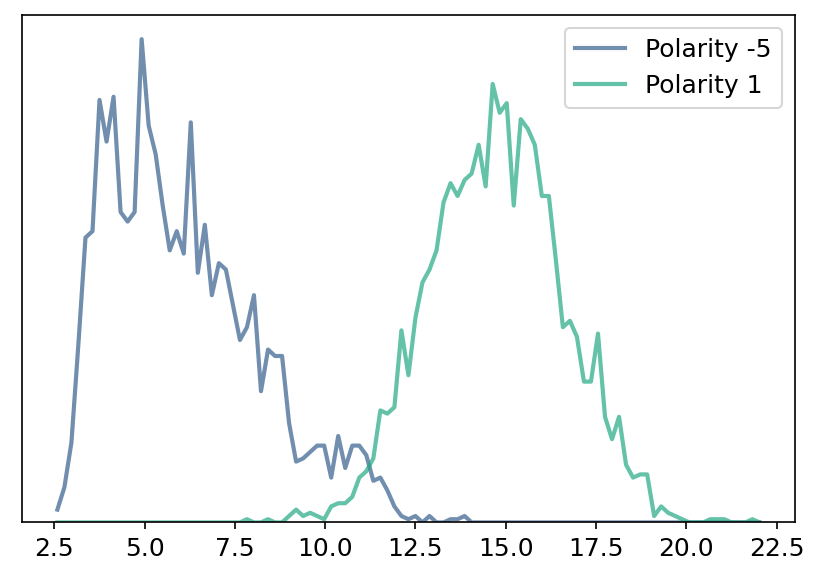}\\
    \end{minipage}
    \begin{minipage}{0.49\linewidth}
    \centering
    \includegraphics[width=\linewidth]{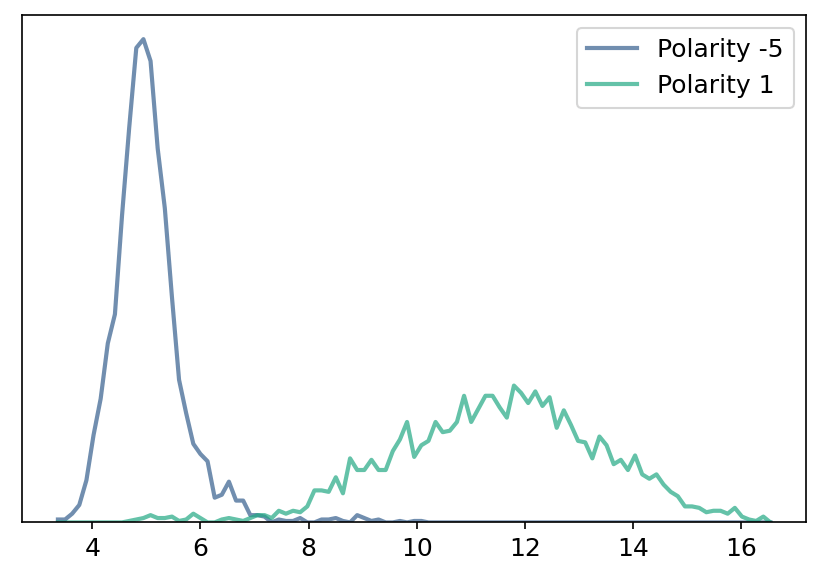}\\
    \end{minipage}
    \vspace{-0.2cm}
    \caption{Distribution of $l_2$ distance to 3 MNIST training set nearest neighbors, for 1000 generated MNIST samples from WGAN {\bf (left)} and NVAE {\bf (right)}. For $\rho=-5$ we see that both distributions have a peak around $5$. For WGAN the distribution has a significantly longer tail compared to NVAE, indicating that the WGAN modes don't necessarily coincide with training points.}
    \label{fig:nn_hist_mnist}
\end{figure}

A last benefit of polarity sampling is to adapt a sampling distribution to a reference distribution that suffered a distribution shift. For example, this could occur when training a DGN on a training set, and using it for content generation with a slightly different type target samples. In fact, as long as the distribution shift remains reachable by the model i.e. in the support of $p_{G}$, altering the value of $\rho$ will help to shift the sampling distribution, possible to better match the target one. In all generality, there is no guarantee that any benefit would happen for $\rho \not = 0$, however, for the particular case where the distribution shift only changes the way the samples are distributed (on the same domain), we observe in \cref{fig:nvae} that $\rho \not = 0$ can provide benefit. To control the experimental setting, we took the colored-MNIST dataset and the NVAE DGN model \cite{vahdat2020nvae} and produce a training set with a Gaussian hue distribution favoring blue
and two test set, one with same hue distribution and one with uniform hue distribution. We observe that $\rho$ can provide a beneficial distribution shift to go from the biased-hue samples to the uniform-hue one.

\subsection{ProGAN Polarity Sweep}
\label{sec:progan}

Previously in \cref{sec:pareto}, we have drawn note to the fact that ProGAN \cite{karras2017progressive}, an architecture which is widely used, but does not incorporate truncation, can also be controlled via polarity sampling. In \cref{tab:progan} we present precision-recall characteristics for polarity sweep on ProGAN. As control, we also perform latent space truncation as in \cite{brock2018large} by sampling a truncated gaussian distribution, parameterized by its support $[-\beta,\beta]$. We change $\beta$ between $[10^{-10},10]$ and notice that for $\beta$ smaller than $10^{-4}$, the generator collapses to 0 precision and recall. Other than that, it maintains a precision of $0.72$ and recall of $0.34$. Using polarity sweep, we also exceed the baseline FID on CelebAHQ 1024x1024 attained by ProGAN; polarity of $-.01$ in pixel-space reduces the FID from $7.37$ to $7.28$.

\subsection{FID for truncated models}
While the FID improvement for some of the methods we present are not significantly large, we see that for truncated models, i.e., models with $\psi<1$, $\rho>0$ provides significant FID improvements by increasing diversity, e.g., for StyleGAN2-FFHQ with $\psi=\{.9,.7,.5\}$, increasing $\rho>0$ improves FID by $\{.69, 8.11, 11.1\}$ points. Such truncation is commonly used in practice for qualitative experiments, making polarity sampling particularly relevant in such settings. Since truncation reduces the range of the generator and polarity increases the diversity of sampling within the range, both can be combined to achieve greater FID improvements.

\subsection{NVAE Negative Log-Likelihood for Varying $\rho$}
To validate the effect of $\rho$ on the likelihood of generated samples, we estimate the negative log-likelihood of samples generated via an NVAE trained on colored-MNIST while varying $\rho$. We generate $5000$ samples each for $\rho=  \{-5,-1,-.5,-.1,0,.1,.5,1,5\}$ which yields negative log-likelihood values of $\{3.0, 3.2, 3.5, 3.8, 3.9, 4.1, 4.3, 4.6, 5.1\}\tiny{{\times}10^{-2}}$ bits/dim. This shows that decreasing $\rho<0$ samples high-likelihood points while increasing $\rho>0$ samples lower-likelihood points compared to standard sampling ($\rho=0$).

\subsection{Perceptual Path Length for Constant Latent Shifts}
\label{appendix:ppl_fixed}
In Sec.~\ref{sec:PPL} we present the PPL variation for a $10^{-4}$ interpolation step from a latent space point towards another random latent space point. To evaluate the perceptual smoothness around regions of the latent space, we also calculate the PPL for paths of length $.3$ starting from individual latent space points towards random directions. We see that for both StyleGAN2-FFHQ and BigGAN-Imagenet, PPL decreases monotonically with $\rho<0$ whereas first decreases and then increases for $\psi<1$. For StyleGAN2-FFHQ we acquire PPL = $\{281,316\}$ for $\psi = \{0.1,0\}$ and PPL = $\{274,271\}$ for $\rho=\{-2,-10\}$. For BigGAN-GoldenRetreiver we get PPL $35.9$ for $\psi=0$ and $0.20$ for $\rho=-2$. This possibly indicates that decreasing truncation might not always lead to perceptually smoother regions whereas decreasing polarity does. 
\begin{table}[]
\centering
\resizebox{\linewidth}{!}{%
\begin{tabular}{@{}ccccccc@{}}
\toprule
 & \multicolumn{6}{r}{CelebAHQ 1024$\times$1024} \\ \midrule
\multicolumn{1}{l}{} & \multicolumn{3}{c}{$\rho \leq 0$} & \multicolumn{3}{c}{$\rho > 0$} \\
$|\rho|$ & FID & Precision & \multicolumn{1}{c|}{Recall} & FID & Precision & Recall \\ \midrule
0 & 7.37 & .73 & .34 & - & - & - \\
0.01 & \textbf{7.28} & .73 & .34 & 7.45 & .73 & .35 \\
0.1 & 7.41 & .76 & .31 & 8.95 & .68 & .38 \\
1 & 12.65 & .85 & .19 & 17.96 & .58 & .48 \\
2 & 13.09 & \textbf{.86} & .19 & 18.54 & .58 & \textbf{.48} \\
\hline
\end{tabular}
}
\caption{FID, Precision and Recall metrics of ProGAN \cite{karras2017progressive} with polarity sweep in the pixel space.}
\label{tab:progan}
\end{table}

\section{Dataset Description}
\label{sec:datasets}

\subsection{colored-MNIST}
\label{sec:hue_dataset}
We perform controlled experiments on NVAE \cite{vahdat2020nvae} by training on datasets with and without controllable distribution shifts. To control the shift, we colorize MNIST with hue ranging $[0,\pi]$ by 1) uniformly sampling the hue and 2) sampling the hue for each image from a truncated normal distribution, with a truncation scale of $2$.

\subsection{LSUN Dataset}
We use the LSUN dataset \cite{yu2015lsun} available at the official website\footnote{https://www.yf.io/p/lsun}. We preprocess the dataset using the StyleGAN2 repository\footnote{https://github.com/NVlabs/stylegan2}. 

\subsection{AFHQv2 and FFHQ}
We use the version 2 of AFHQ that was released prepackaged with StyleGAN3 for our experiments. For FFHQ we use also use TFRecords provided with StyleGAN2.  

\subsection{License}
The majority of Polarity-Sampling is licensed under CC-BY-NC, however portions of the project are available under separate license terms: NVAE, StyleGAN2 and StyleGAN3 are licensed under the NVIDIA license; guided-diffusion is licensed under the MIT license.\footnote{https://github.com/openai/guided-diffusion/blob/main/LICENSE}

\section{Qualitative results}
\label{sec:quality}

\begin{figure*}[t!]
\centering
\begin{center}
\begin{minipage}{0.97\textwidth}
\centering
      \includegraphics[width=.86\linewidth]{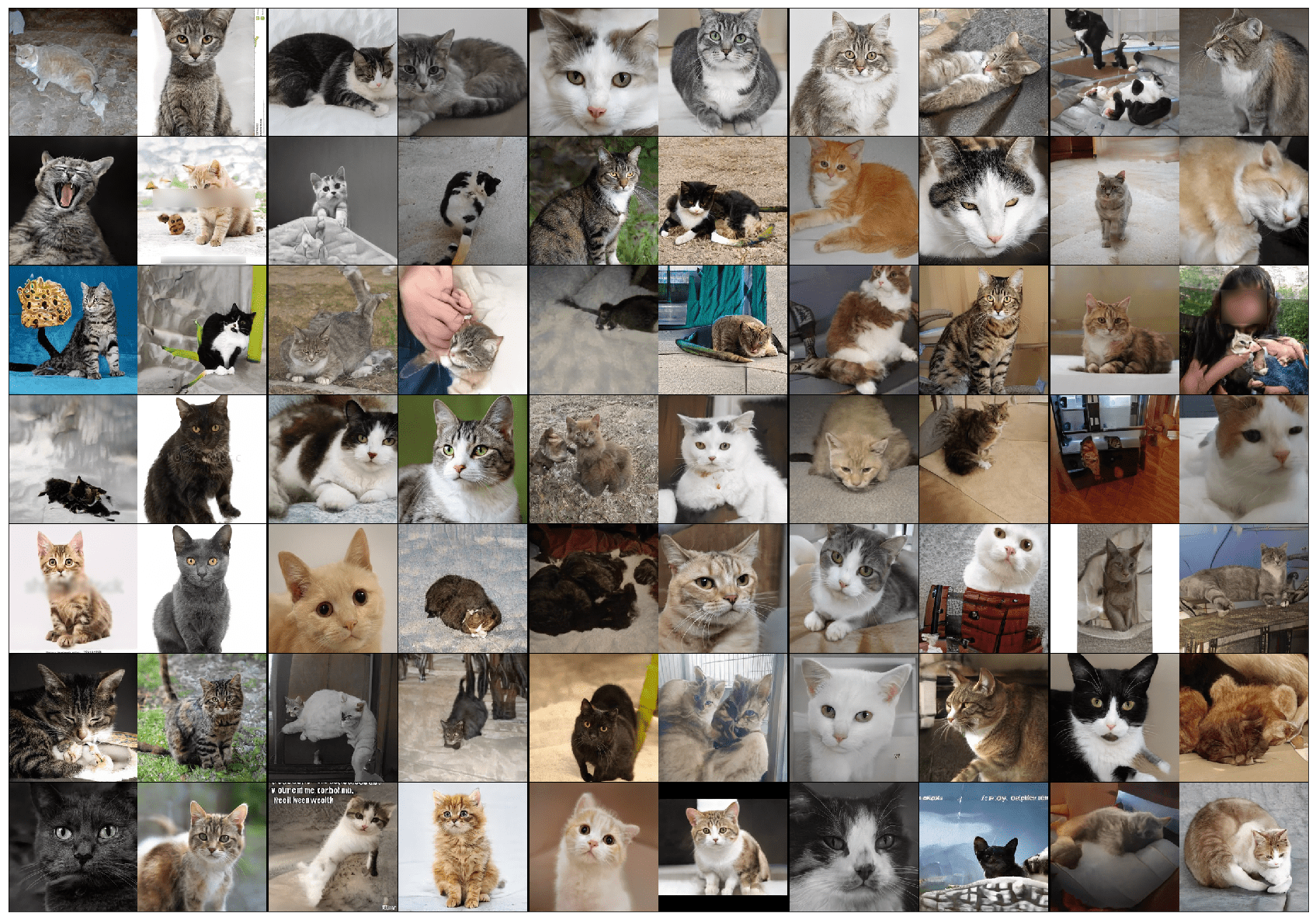}
\end{minipage}\\
\begin{minipage}{0.97\textwidth}
\centering
      \includegraphics[width=.86\linewidth]{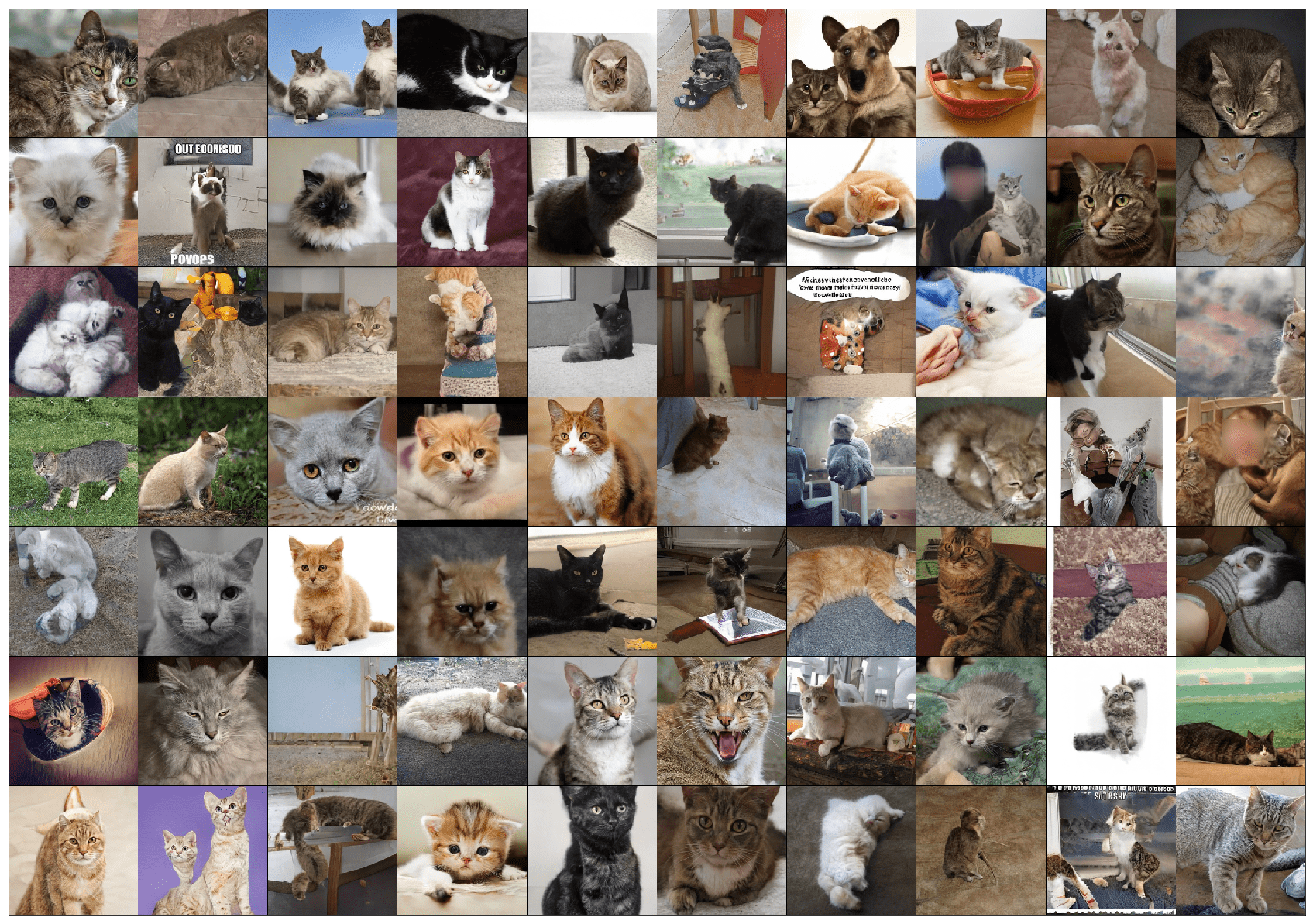}
\end{minipage}    
\end{center}
\vspace{-0.2cm}
\caption{Uncurated samples of LSUN Cats using (top)  $\rho=\{-1,-.5,-.2,-.1,0\}, \psi=.8$ and (bottom) $\psi=\{.7,.73,.75,.77,.8\}$; both representing regions with roughly an equal span of recall score on \cref{fig:pareto}. Notice the significant precision of the left-most columns of top compared to the left-most of bottom, where at equal diversity, top has significantly higher precision score.}
\label{fig:lsun_cat_equilen}
\end{figure*}

\begin{figure*}[t!]
\centering
\begin{center}
\begin{minipage}{\textwidth}
      \includegraphics[width=\linewidth]{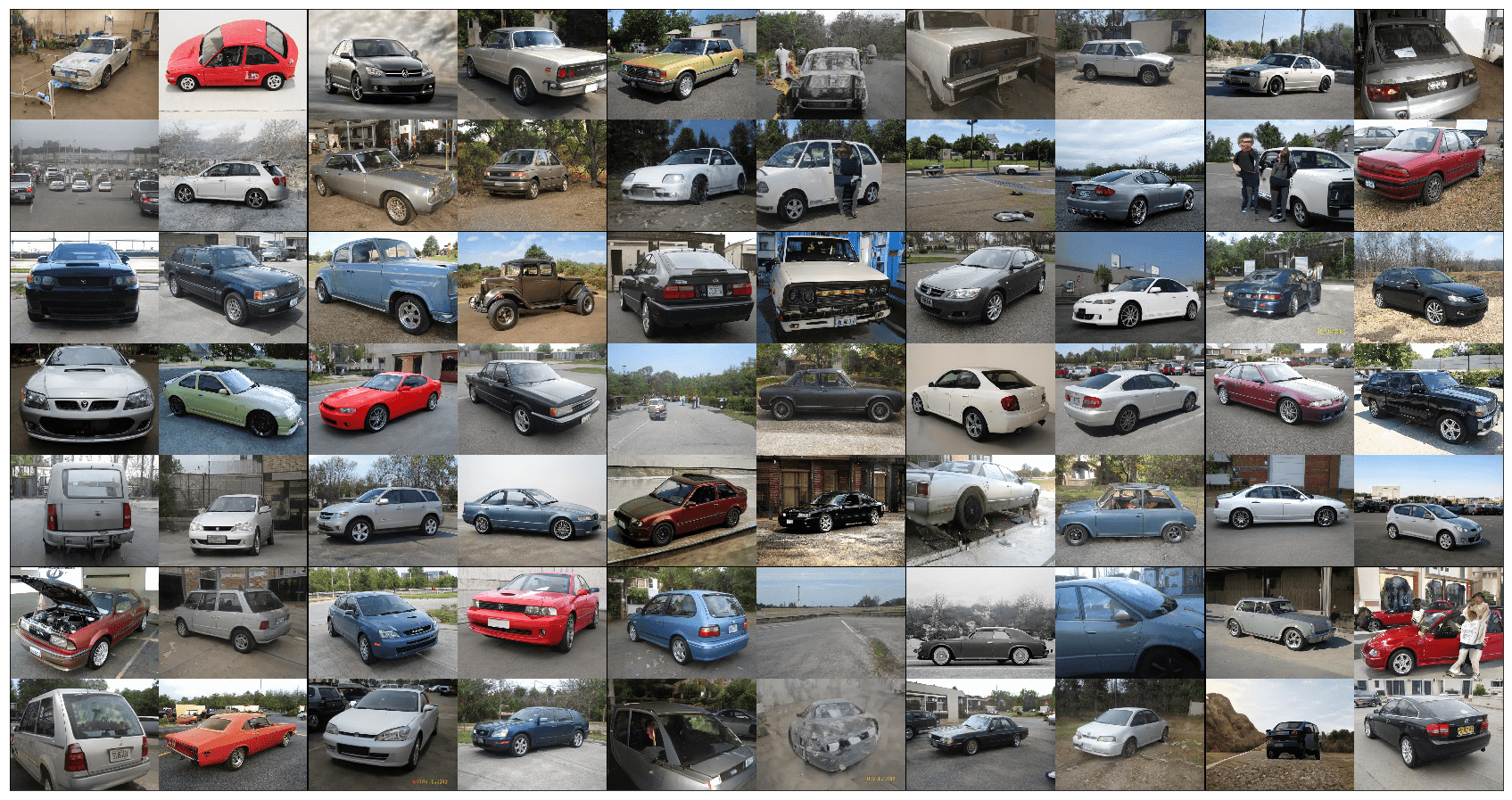}
\end{minipage}\\
\begin{minipage}{\textwidth}
      \includegraphics[width=\linewidth]{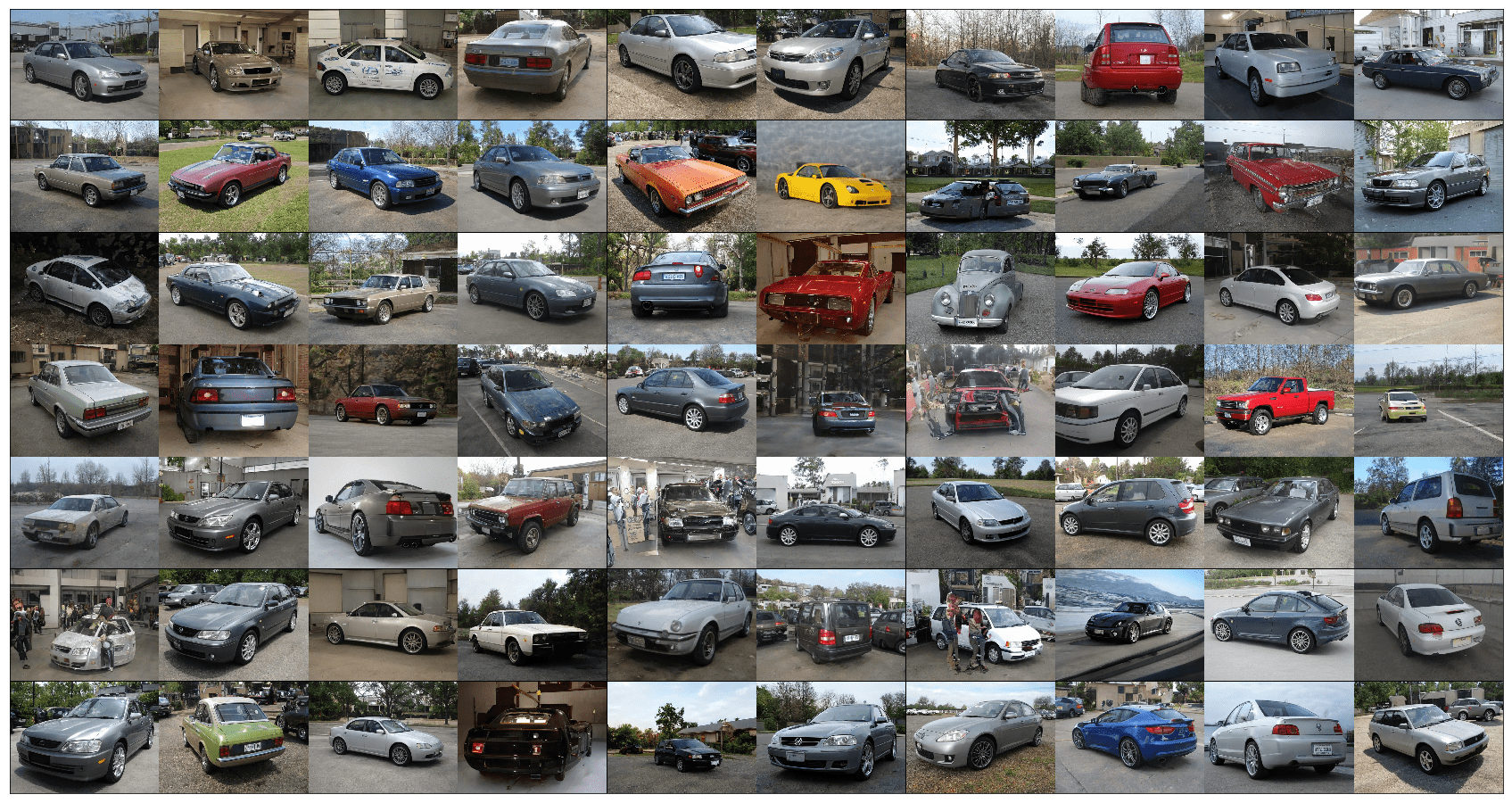}
\end{minipage}    
\end{center}

\vspace{-0.2cm}
\caption{Uncurated samples of LSUN Cars using (top)  $\rho=\{-.1,-.075,-.05,-.025,0\}, \psi=.7$ and (bottom) $\psi=\{.5,.55,.6,.65,.7\}$; both representing regions with roughly an equal span of precision score on \cref{fig:pareto} Notice the significant diversity in (top) especially in the leftmost columns, where the recall score is significantly higher than that of the leftmost column of bottom left, with equal precision.
}
\label{fig:qualitative_alpha_sweep_cars}
\end{figure*}

\begin{figure*}[t!]
\centering
\begin{center}
\begin{minipage}{0.97\textwidth}
\centering
      \includegraphics[width=0.93\linewidth]{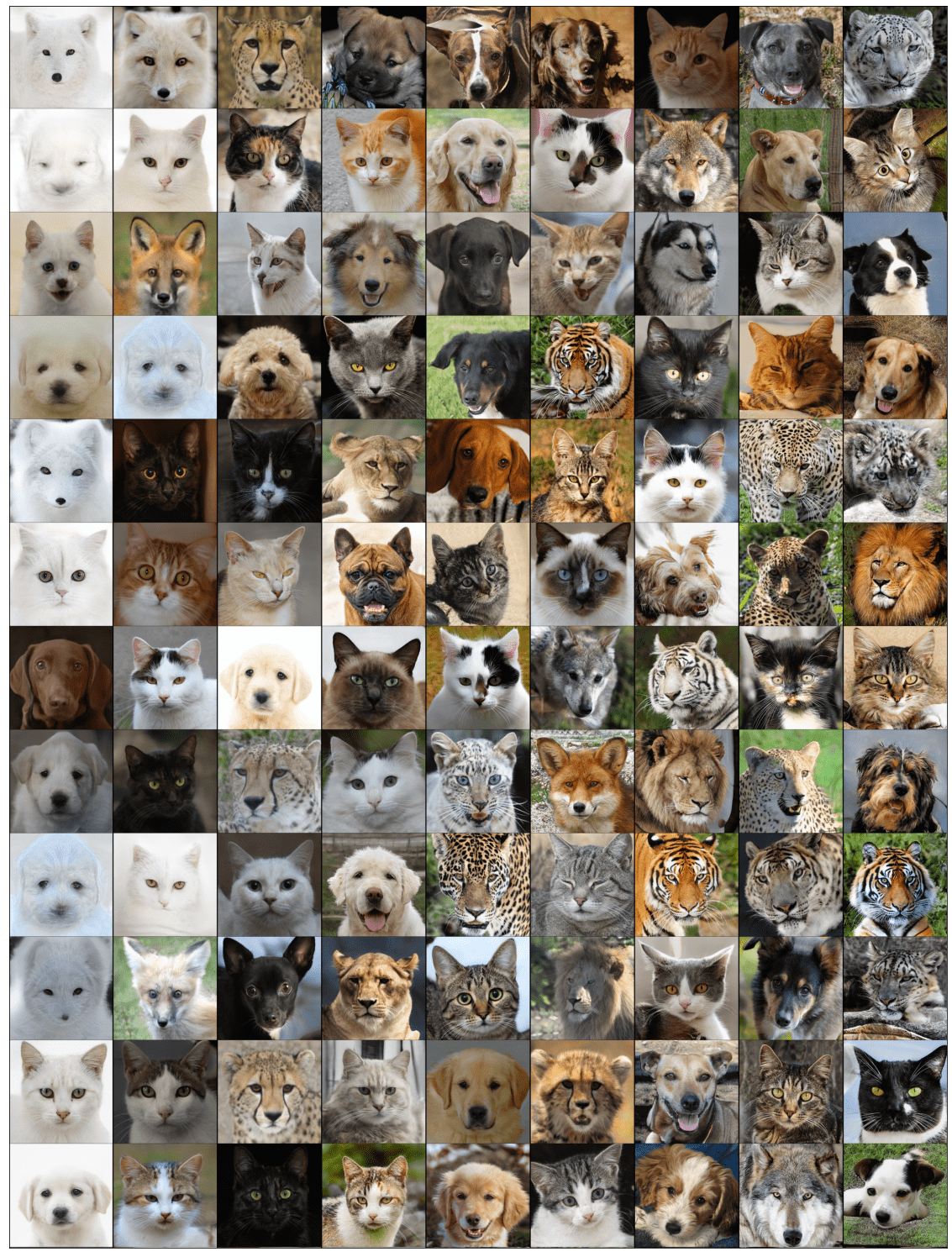}
\end{minipage}    
\end{center}

\vspace{-0.2cm}
\caption{Uncurated samples of AFHQv2 using $\rho=\{-2,-1,-.5,-.2,0,.01,.1,.2,.5\}$ and $\psi=.9$ in pixel-space. As we move right from baseline (middle column) we see an increase in texture diversity of images, whereas, moving left, we see images with smoother textures. 
}
\label{fig:qualitative_alpha_sweep_afhq}
\end{figure*}

\begin{figure*}[t!]
    \centering
    \includegraphics[width=0.8\linewidth]{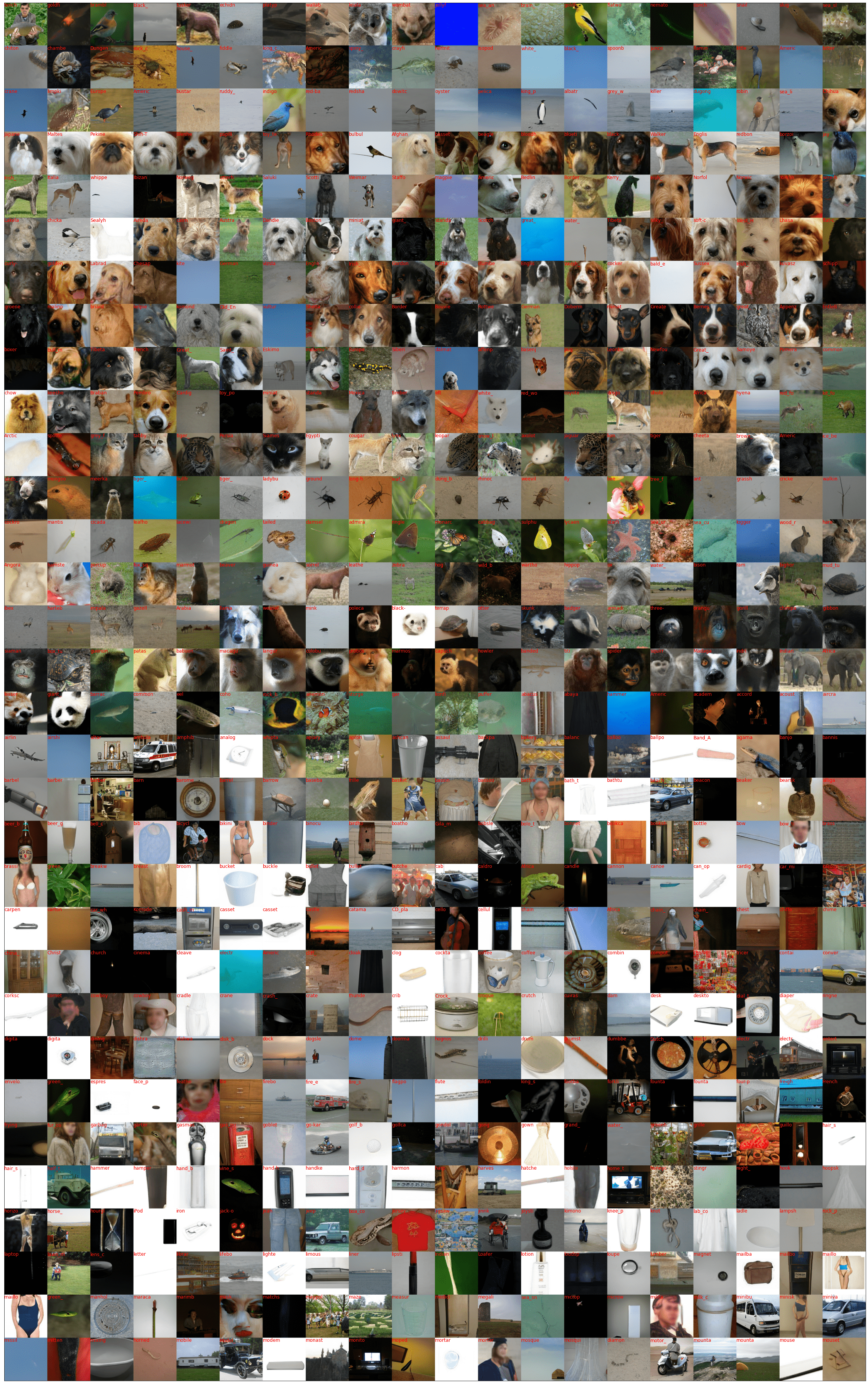}
    \caption{Depiction of a single mode (large negative $\rho$) for each class of the first $800$ Imagenet classes.}
    \label{fig:modes_per_class}
\end{figure*}

We provide in the following pages,  \cref{fig:lsun_cat_equilen,fig:qualitative_alpha_sweep_cars,fig:qualitative_alpha_sweep_afhq,fig:modes_per_class} that correspond to LSUN Cats, LSUN Cars, AFHQv2 samples with varying $\rho$ values and $800$ Imagenet modes. For LSUN Cars and LSUN Cats we draw comparisons between varying truncation and varying polarity independently. 

\end{document}